\documentclass[letterpaper]{article} 
\usepackage{aaai25}  
\usepackage{times}  
\usepackage{helvet}  
\usepackage{courier}  
\usepackage[hyphens]{url}  
\usepackage{graphicx} 
\usepackage{natbib}  
\usepackage{caption} 
\usepackage{algorithm}
\usepackage{algorithmic}
\usepackage{subcaption}
\usepackage{graphicx}
\usepackage{multirow}

\usepackage{amssymb}
\usepackage{amsmath}

\usepackage{newfloat}
\usepackage{listings}
\DeclareCaptionStyle{ruled}{labelfont=normalfont,labelsep=colon,strut=off} 
\floatstyle{ruled}
\newfloat{listing}{tb}{lst}{}
\floatname{listing}{Listing}
\usepackage{color}

\title{Performance Optimization of Ratings-Based Reinforcement Learning}
\author {
    Evelyn Rose\textsuperscript{\rm 1},
    Devin White\textsuperscript{\rm 2},
    Mingkang Wu\textsuperscript{\rm 1},
    Vernon Lawhern\textsuperscript{\rm 3},
    Nicholas R. Waytowich\textsuperscript{\rm 3},\\
    Yongcan Cao\textsuperscript{\rm 1}
}
\affiliations {
    \textsuperscript{\rm 1}The University of Texas at San Antonio\\
    \textsuperscript{\rm 2}Army Educational Outreach Program\\
    \textsuperscript{\rm 3}DEVCOM Army Research Lab
}

\begin{document}

\maketitle
\begin{abstract}
This paper explores multiple optimization methods to improve the performance of rating-based reinforcement learning (RbRL). RbRL, a method based on the idea of human ratings, has been developed to infer reward functions in reward-free environments for the subsequent policy learning via standard reinforcement learning, which requires the availability of reward functions. Specifically, RbRL minimizes the cross entropy loss that quantifies the differences between human ratings and estimated ratings derived from the inferred reward. Hence, a low loss means a high degree of consistency between human ratings and estimated ratings. Despite its simple form, RbRL has various hyperparameters and can be sensitive to various factors. Therefore, it is critical to provide comprehensive experiments to understand the impact of various hyperparameters on the performance of RbRL. This paper is a work in progress, providing users some general guidelines on how to select hyperparameters in RbRL. 


\end{abstract}

\section{Introduction}
\label{sec:intro}

Recent advancement in reinforcement learning (RL) has shown its potential to solve a wide range of tasks from mastering Atari games \cite{badia2020agent57outperformingatarihuman} to complex robotics tasks \cite{tang2024deepreinforcementlearningrobotics}. One key requirement when employing the developed RL algorithms is the existence of properly defined reward functions. However, the design of proper reward functions for real world tasks is often difficult or even infeasible due to the large state and action spaces. Often, a slightly different reward function can lead to a completely different policy even if trained via the same RL algorithm. Hence, it is critical to investigate how to infer reward functions without manually assigning reward functions. 

One typical approach to infer reward function is RL from human feedback (RLHF), which seeks to solicit human feedback that can be used to learn a reward signal. The main principle of RLHF is that the learned reward can explain the human feedback well. In other words, the estimated feedback from the learned reward should match the actual human feedback. 
Preference-based reinforcement learning (PbRL) \cite{christiano2017deep} is a well-know RLHF method that utilizes humans' preferences over sample pairs to learn the reward function. A well trained reward function can yield estimated preferences that match humans' preference well. In other words, if one segment is preferred over the other one in the pair by the human user, the cumulative reward of the former based on the learned reward should be larger than that of the later. Recently, rating-based reinforcement Learning (RbRL) \cite{white2024rating}, was developed that utilizes humans' ratings of individual samples to learn the reward function. It is shown in \cite{white2024rating} that ratings can be easy and more acceptable to humans as comparing two similar samples in PbRL can be challenging. Similar to PbRL, a well trained reward function can yield estimated ratings that match humans' ratings. In other words, if one segment has a high rating, its cumulative reward should be relatively high. However, for two samples with the same ratings, their cumulative rewards can still be different. The allowable differences among samples in one rating depends on the human user's standard, which can be different from one to another. In addition to the different types of human feedback, RbRL has more hyperparameters than PbRL, which makes it more challenging yet important to optimize RbRL.   

In this paper, we focus on addressing the gap of optimizing RbRL via leveraging some key principles in the existing machine learning literature. For example, existing studies on machine learing optimization include the use of different optimizers\cite{loshchilov2019decoupledweightdecayregularization}, the inclusion of dropout \cite{srivastava2014dropout}, the use of different number of hidden layers \cite{4196403}, and different activation functions and learning rates \textcolor{black}{\cite{wu2019demystifyinglearningratepolicies, dubey2022activationfunctionsdeeplearning}}. Moreover, we will also investigate other optimization techniques unique to RbRL, including, e.g., design of rating probability estimation functions, reward boundary selections, and confidence index $k$, to provide a comprehensive study on how the hyperparameters of RbRL should be selected for performance optimization. 

In this paper, we will perform a \textcolor{black}{preliminary set of experiments} aimed at identifying the best optimization techniques to improve both consistency and performance of RbRL. We perform a total of 8 optimization techniques, including classic machine learning optimization techniques and optimization methods that are unique to RbRL. Our experiments show that by performing standard optimization techniques in some cases, we can drastically improve the performance by 100\%. 
Meanwhile, we show that the optimized RbRL can retain a consistent performance across different numbers of rating classes with much less variations. 
\textcolor{black}{The optimization methods used in this paper includes: (1) reward boundary, (2) confidence index, (3) novel class probability function, (4) activation functions, (5) learning rate, (6) AdamW vs. Adam optimizer, (7) number of hidden layers, and (8)) dropout rate. }
The first three optimization methods are unique to RbRL while the remaining five are common machine learning optimization methods.

\section{Intersection of Collaborative AI and Modeling Humans}

\begin{figure}[h!]
    \centering
    \includegraphics[width=\linewidth]{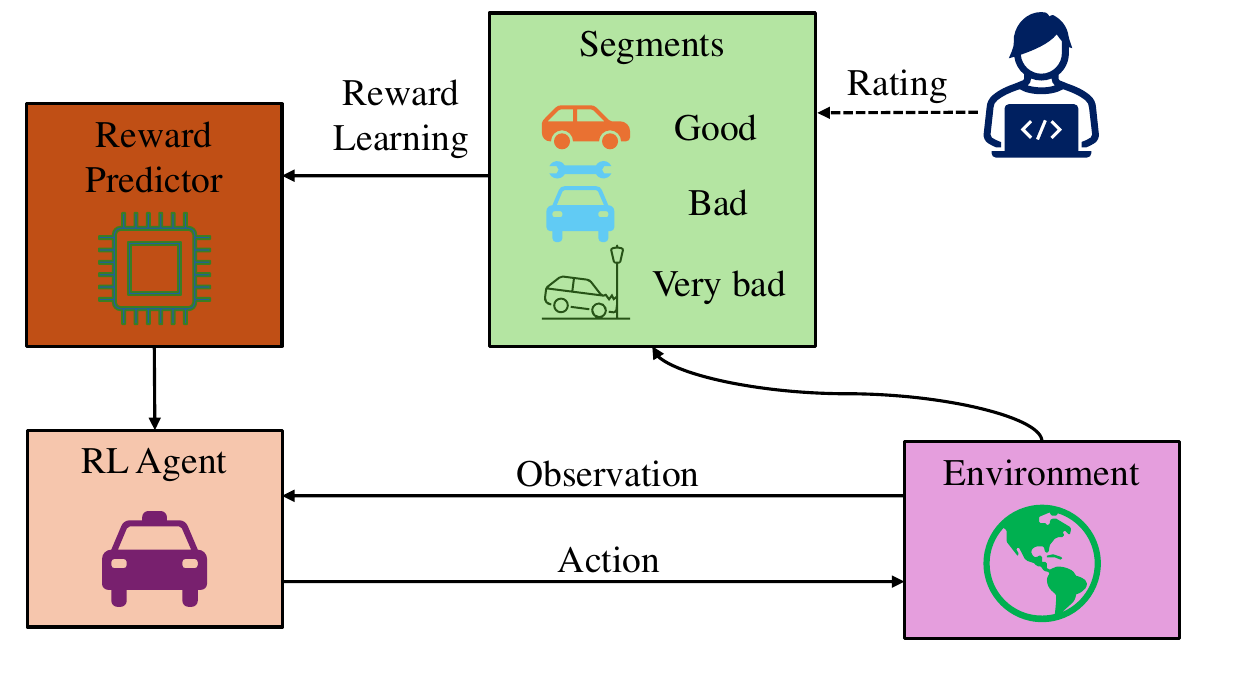}
    \caption{A schematic illustration of rating-based reinforcement learning.}
    \label{fig:flowchart}
\end{figure}

RbRL explores learning policy via reinforcement learning from human ratings with the scheme shown in Figure \ref{fig:flowchart}. Within the RL framework, different reward assignment strategies offer different interpretations of human ratings. This aligns with the way humans employ diverse strategies to achieve goals, progressing from lower-rated to higher-rated examples as guidance. In terms of RL, leveraging rewards based on human ratings fosters a more human-like learning process, distinct from conventional machine-driven approaches which relies on uniform reward assignment. From a cognitive standpoint, RL expands the range of examples beyond the familiar scenarios humans typically consider, offering more diverse segments for ratings and enabling more accurate reward inference.

Our proposed optimizations aim to help foster a more collaborative modeling between RL and a user's inherent reward signal. By investigating the optimization of RbRL, we will investigate how we can further improve this map of a users reward signal for a task to the RL algorithm.

\section{Related Works}
\label{sec:related_works}
The absence of rewards is a crucial challenge in RL. Many approaches tackle this issue by designing specific reward functions for each environment. Methods such as inverse reinforcement learning (IRL) \cite{ng2000algorithms, ziebart2008maximum, finn2016guided, levine2011nonlinear, choi2011map, szot2022bc} derive reward functions from expert demonstrations to guide policy updating. However, these methods heavily depend on the quality of demonstrations, making them challenging to implement due to the high cost and difficulty in obtaining expert demonstrations.

Recently, inferring reward functions from human guidance has emerged as an effective alternative to relying on expert demonstrations. Approaches such as PbRL and RbRL \cite{christiano2017deep, white2024rating} require only a basic understanding of the environment from human users to provide meaningful guidance. PbRL infers reward functions based on human preferences between two segments sampled from the current policy. By distinguishing between these segments, a reward function is trained to mimic the decision-making process behind human preferences. RbRL enhances PbRL by introducing a multi-level rating mechanism for segments sampled from the current policy. Instead of relying on pairwise preferences, RbRL assigns ratings to individual segments, enabling the reward function to be trained to capture human evaluations for each state-action pair more effectively, leading to improved performance. Typically, reward functions are modeled using a classification approach to distinguish between different samples. As a result, the quality of the trained reward function directly impacts performance, meaning, a well-trained model yields better outcomes, and vice versa.

However, it can be observed that, in some cases, the performance of RbRL decreases as the number of rating classes increases. This issue arises from the imbalanced data distribution across the rating classes, as the rating process in RbRL typically occurs at the beginning of training, when the random policy generates segments of uneven quality, leading to an unequal distribution of segments across the rating classes. Thereby, our work aims to address the issue of imbalanced rating distributions in RbRL without changing its framework. Instead, we explore various optimization methods to refine the classification part of reward learning in RbRL.

\section{Problem Statement}\label{preliminaries}
\subsection{Problem Formulation}

We frame our problem as a Markov Decision Process (MDP) with human ratings in a reward-free setting. This process can be represented by a tuple $(S, A, P, \gamma, n)$, where $S$ is the state space, $A$ is the action space, $P$ denotes the state transition probability distribution, $\gamma$ is the discount factor mitigating the impact of infinite rewards in standard RL, and $n$ is the number of rating classes. For each state $s \in S$, the RL agent takes an action $a \in A$ sampled by the RL policy $\pi$. The goal of RL is to find an optimal policy $\pi^*$ that maximizes the cumulative rewards of the agent's behaviors. In standard RL, this process can be formulated as
\begin{equation} \label{eq: std RL goal}
    \pi^* = \arg\max_{\pi}\sum_{t=0}^{\infty} \mathbb{E}_{(s_t, a_t) \sim \rho_{\pi}} \Big [ \gamma^t r(s_t, a_t)\Big],
\end{equation}
where $ \rho_{\pi} $ is a marginal state-action distribution given a policy $\pi$. In standard RL, the reward $r$ is generated from the reward function $R$ which is provided by the environment. In RbRL framework where rewards are unavailable, a reward function $\hat{R}$ is constructed using $n$ classes of human ratings. This learned reward function generates the reward $\hat{r}$, which is then used to update the policy. Consequently, in RbRL, \eqref{eq: std RL goal} can be rewritten as
\begin{equation} \label{eq: rbrl goal}
    \pi^* = \arg\max_{\pi}\sum_{t=0}^{\infty} \mathbb{E}_{(s_t, a_t) \sim \rho_{\pi}} \Big[ \gamma^t \hat{r}(s_t, a_t)\Big].
\end{equation}

The specific methodology for learning the reward function will be detailed in the following sections.

\subsection{Rating-based Reinforcement Learning~\cite{white2024rating}}


To learn the reward from human ratings, let's define a segment (short video clip) as a set of state action pairs $\sigma = ((s_0, a_0), (s_1, a_1), (s_2, a_2), ..., (s_n, a_n))$. This segment is then mapped to an inferred reward $\hat{r}(s,a)$ via a reward predictor. To do this, the first step is to define cumulative reward for a segment $\hat{R}(\sigma) = \sum_{i=0}^{n} \hat{r}(s_i, a_i)$ and is then normalized across a batch as $$\tilde{R}(\sigma) = \frac{\hat{R}(\sigma) - \min_{\sigma' \in X} \hat{R}(\sigma')} {\max_{\sigma' \in X}\hat{R}(\sigma') - \min_{\sigma' \in X} \hat{R}(\sigma')},$$
where $X$ is the samples in a batch.
Since the number of samples in a given class for the batch is known, we can find a lower and upper bound reward for a given class such that the total number of samples within the lower and upper bounds matches the number of samples in that class based on human ratings. By defining the lower and lower bound of a class $i$ as $\bar{R}_i$ and $\bar{R}_{i+1}$, the predicted probability for a sample in the rating class $i$ is given by
\begin{equation}\label{eq:Qfun_old}
     Q_{\sigma}(i) = \frac{e^{-k(\tilde{R}(\sigma)-\bar{R}_i)(\tilde{R}(\sigma)-\bar{R}_{i + 1})}}{\sum_{j=0}^{n-1} e^{-k(\tilde{R}(\sigma)-\bar{R}_j)(\tilde{R}(\sigma)-\bar{R}_{j + 1})}},
\end{equation}
where $k$ is the confidence index indicating the user's relative confidence. A higher $k$ means a high confidence, vice versa.
The reward predictor then infers the reward function via minimizing the cross entropy loss given by 
\begin{equation}\label{eq:crossentropy}
    L(\hat{r}) = -\sum_{\sigma\in X} \left( \sum_{i=0}^{n-1}\mu_{\sigma}(i) \log \big(Q_{\sigma}(i)\big) \right),
\end{equation}
where $\mu_{\sigma}(i) =1$ if the sample $\sigma$ is rated in class $i$ by the user and $\mu_{\sigma}(i) =0$ otherwise. 

Once the reward function is learned, many existing RL algorithms, such as PPO, A2C, SAC, can be used to train policies based on the learned reward function.

\subsection{Optimization}

Let's now focus on the optimization method to be performed on the RbRL algorithm. To simplify our studies, we have used synthetic humans in our studies, where synthetic ratings are generated based on the true environment reward. For example, if the maximum true environment reward for a sample is $50$ and the number of rating classes is $2$, the threshold between rating class $0$ (``Bad") and rating class $1$ (``Good") can be any value between $0$ and $50$. A higher threshold means a higher bar for the samples in rating class $1$ and vice versa. Our first optimization method is to quantify the impact of the reward boundary on the performance of RbRL. 

The second optimization method is the design of a different function that predicts the probability of a sample in the rating class $i$. In the existing approach~\eqref{eq:Qfun_old} proposed in~\cite{white2024rating}, the probability is determined based on the distances between the cumulative reward of the current sample and the lower/upper bounds. Different from this method, we here propose a new form that is based on the distance between the cumulative reward of the current sample and the average of the upper and lower bounds. In other words, the new method is a direct measure of the deviation of the cumulative reward of the current sample from the middle of the rating class $i$. Specifically, the new predicted probability is given by
\begin{equation}\label{eq:OtherQ}
     Q_{\sigma}(i) = \frac{e^{-k\left(\tilde{R}(\sigma)-\frac{\bar{R}_{i}+\bar{R}_{i + 1}}{2}\right)^2}}{\sum_{j=0}^{n-1} e^{-k\left(\tilde{R}(\sigma)-\frac{\bar{R}_{j}+\bar{R}_{j + 1}}{2}\right)^2}}.
\end{equation}
This method is referred to as novel class probability function optimization.

Additionally, we will also quantify how the confidence index $k$ in \eqref{eq:Qfun_old} may impact the performance of RbRL, which is the third method to b considered in this work. These three methods are unique to the RbRL method.


In addition to these three optimization methods, we also investigate a wide range of traditional optimization methods from classic machine learning. For example, we will perform a comprehensive set of tests on performance evaluation via (1) using the AdamW optimizer versus Adam optimizer, (2) different Dropout rates, (3) different number of hidden layers, (4) different activation functions and (5) different learning rate.  

\section{Results}

Our experimental set up utilizes the codebase for Rating-based Reinforcement Learning \cite{white2024rating}, a variation of BPref \cite{lee2021b}. We use Proximal Policy Optimization \cite{schulman2017proximalpolicyoptimizationalgorithms} as our RL algorithm, where we replace the true environment reward with a predicted reward based on the reward predictor. For the first 32,000 timesteps, an entropy based exploration reward is used to gather initial segments which will be rated. After the initial 32,000 timesteps, only the predicted reward is used in learning. In our experiments, a total of 1,000 ratings was used for Walker. 2,000 ratings were used for both Quadruped and Cheetah. 


\subsection{Confidence Index}

We perform tests for different confidence index $k$ to see how both \eqref{eq:Qfun_old} and \eqref{eq:OtherQ} perform with respect to $k$ for Quadruped, Cheetah and Walker, shown in Table \ref{tab:kresults}. It can be seen that for  \eqref{eq:Qfun_old}, $k=1$ yields the highest average performance for $n=2$ and $n=6$ in Walker, the highest average performance for $n=6$ for quadruped and the highest average performance for $n=2$ for cheetah. However, for \eqref{eq:OtherQ}, the results show that $k=0.5$ works best for $n=2$ in Walker and $k=5$ works best for $n=6$ in Walker, while $k=5$ works best for $n=2$ in Quadruped and $k=0.1$ works best for $n=6$ in Quadruped and $k=1$ works best for $n=2$ and $n=6$ in Cheetah. 

Although the optimal $k$ may be different for different $n$ and environments, selecting an unusual small or large $k$ can negatively impact the performance of RbRL. in our tests, $k=1$ can produce best performance in most cases. Hence, it is reasonable to select $k$ near the value of $1$ in practice. 




\begin{table*}[ht!]
  \centering
  \begin{tabular}{p{1.3cm}|p{2.3cm}|p{2.3cm}|p{2.3cm}|p{2.3cm}|p{2.3cm}|p{2.3cm}}
    \hline
    {$k$ in \eqref{eq:Qfun_old}} & \multicolumn{6}{c}{Episodic Reward} \\
    \cline{2-7}
    & Walker RbRL (n=2) & Walker RbRL (n=6) & Quadruped RbRL (n=2) & Quadruped RbRL (n=6) & Cheetah RbRL (n=2) & Cheetah RbRL (n=6)\\
    \hline
    $k = 0.1$ & $889.1 \pm 27.72$          & $915.46 \pm 4.63$          & $132.64 \pm 30.24$        & $228.37 \pm 228.37$         & $444.23 \pm 28.5$  & $403.3 \pm 13.12$\\
    $k = 0.5$ & $890.98 \pm 40.26$         & $915.92 \pm 3.21$          & $178.31 \pm 24.57$        & $370.46 \pm 113.87$         & $371.59 \pm 5.69$  & $\mathbf{453.08 \pm 22.54}$\\
    $k = 1$   & $\mathbf{922.85 \pm 2.88}$ & $\mathbf{931.09 \pm 9.45}$ & $149.26 \pm 27.57$        & $\mathbf{477.96 \pm 137.8}$ & $\mathbf{481.35 \pm 12.65}$ & $412.57 \pm 24.57$\\
    $k = 5$   & $872.56 \pm 1.77$          & $913.56 \pm 5.01$          & $\mathbf{248.4 \pm 92.8}$ & $474.21 \pm 138.28$         & $454.92 \pm 27.17$ & $324.18 \pm 68.98$\\
    $k = 10$  & $839.21 \pm 23.66$         & $912.07 \pm 1.02$          & $197.2 \pm 93.75$         & $380.64 \pm 191.72$         & $426.55 \pm 74.52$ & $420.41 \pm 28.53$\\
    \hline
    {$k$ in \eqref{eq:OtherQ}} \\
    \hline
    $k = 0.1$ & $910.67 \pm 14.73$         & $912.07 \pm 3.17$         & $194.45 \pm 92.98$          & $\mathbf{614.67 \pm 163.7}$ & $354.02 \pm 22.25$ & $415.14 \pm 13.45$ \\
    $k = 0.5$ & $\mathbf{917.17 \pm 13.3}$ & $853.21 \pm 14.65$        & $130.97 \pm 17.28$          & $406.19 \pm 173.37$         & $370.16 \pm 21.58$ & $373.18 \pm 13.56$ \\
    $k = 1$   & $864.39 \pm 21.58$         & $911.83 \pm 18.63$        & $101.86 \pm 23.65$          & $559.54 \pm 192.21$         & $\mathbf{478.69 \pm 84.02}$ & $\mathbf{428.1  \pm 31.33}$ \\
    $k = 5$   & $893.06 \pm 12.11$         & $\mathbf{941.11 \pm 3.9}$ & $\mathbf{217.65 \pm 70.11}$ & $445.93 \pm 154.31$         & $364.99 \pm 26.21$ & $393.07 \pm 60.7$ \\
    $k = 10$  & $846.01 \pm 20.3$          & $927.57 \pm 10.61$        & $192.35 \pm 26.26$          & $195.81 \pm 12.91$          & $393.09 \pm 48.82$ & $397.05 \pm 64.61$ \\
    \hline
\end{tabular}
\caption{Episode reward for various confidence indices in Walker, Quadruped, and Cheetah.}
\label{tab:kresults}
\end{table*}
\subsection{Reward Boundary}

The reward boundary plays an important role in differentiating the segments across different rating classes in both Equations \eqref{eq:Qfun_old} and \eqref{eq:OtherQ}. As our experiment are based on the synthetic data, this reward boundary serves as a proxy for a human rater's criteria. Investigating the impact of varying reward boundaries is essential for understanding how differences in human raters' criteria impact performance.

\begin{figure}[ht!]
   \centering
   \begin{subfigure}[b]{0.49\columnwidth}
        \centering
        \includegraphics[width=\columnwidth]{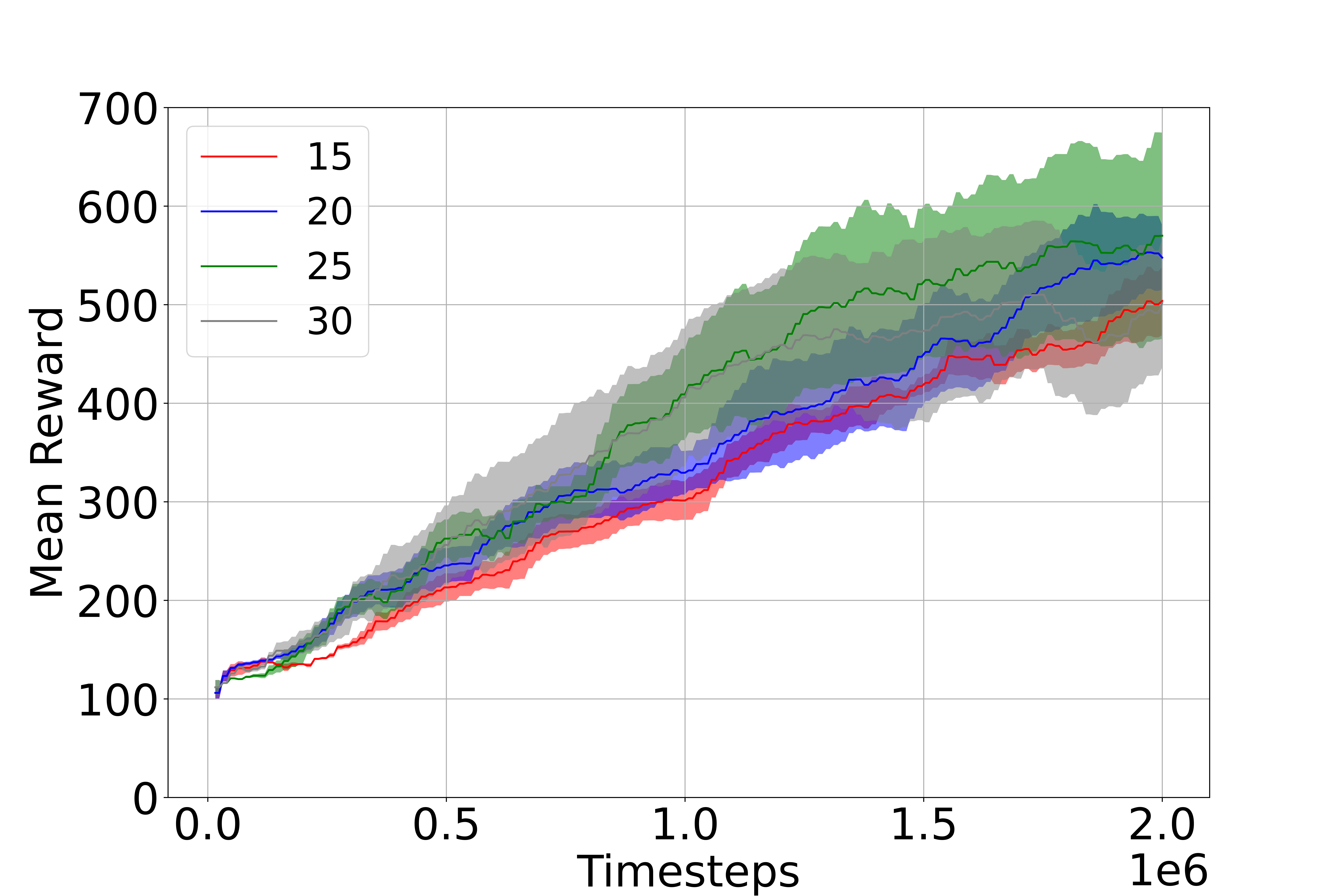}
        \caption{\eqref{eq:Qfun_old} with $n=2$}
        \label{fig:quad_k_value}
   \end{subfigure}
   \begin{subfigure}[b]{0.49\columnwidth}
        \centering
         \includegraphics[width=\columnwidth]{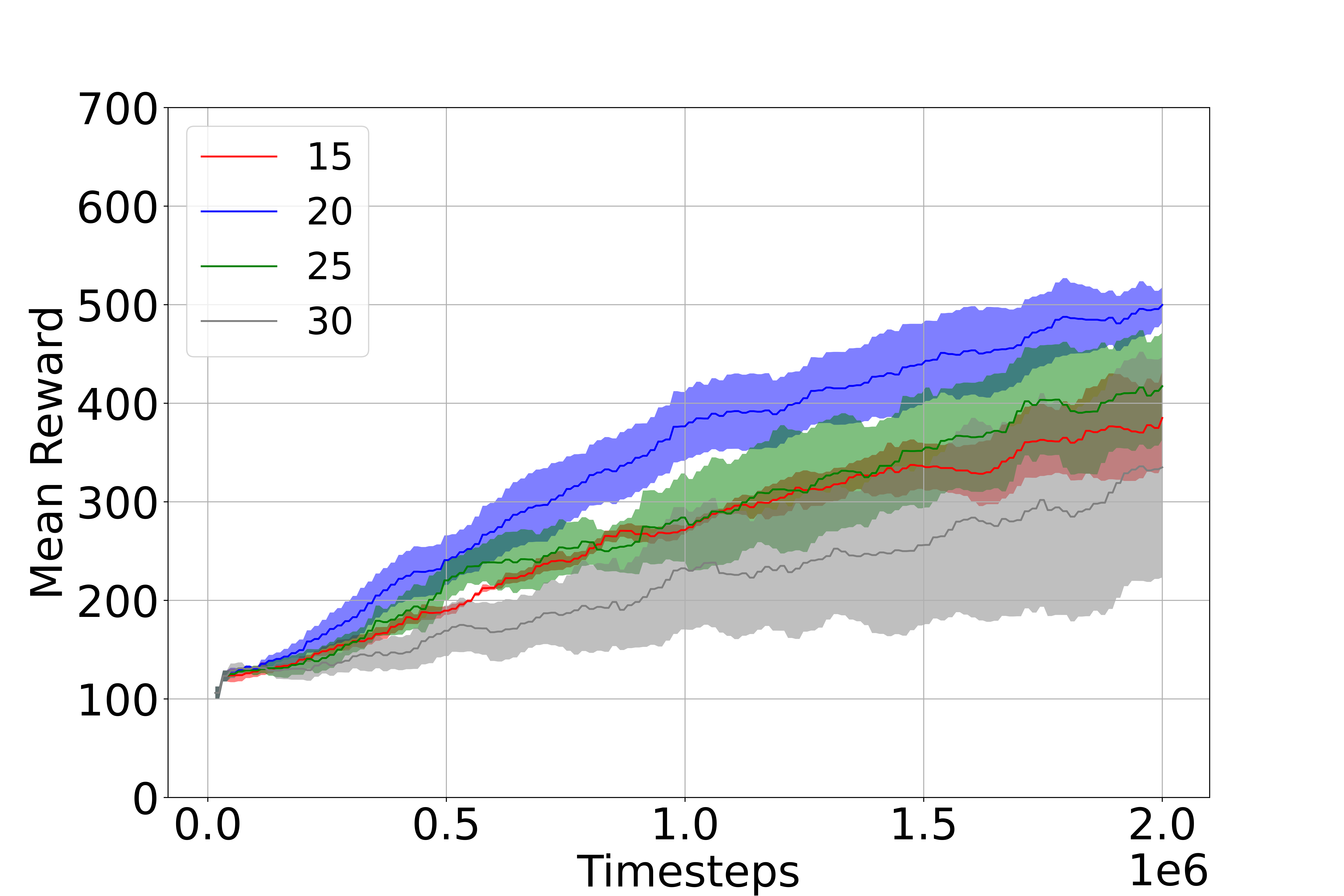}
         \caption{\eqref{eq:Qfun_old} with $n=6$}
         \label{fig:quad_k_value}
   \end{subfigure}
   \begin{subfigure}[b]{0.49\columnwidth}
        \centering
        \includegraphics[width=\columnwidth]{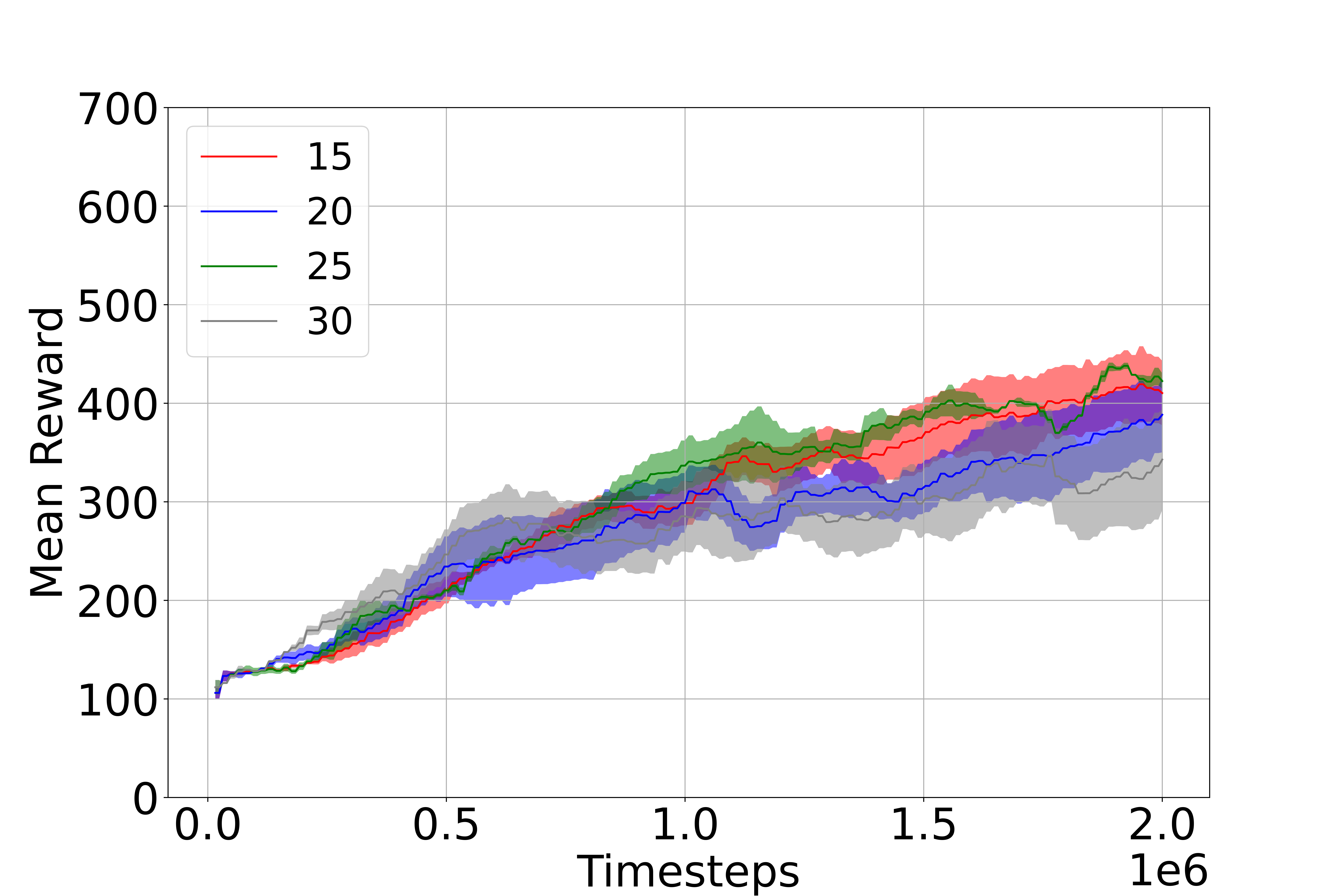}
        \caption{\eqref{eq:OtherQ} with $n=2$}
        \label{fig:quad_k_value}
   \end{subfigure}
   \begin{subfigure}[b]{0.49\columnwidth}
        \centering
        \includegraphics[width=\columnwidth]{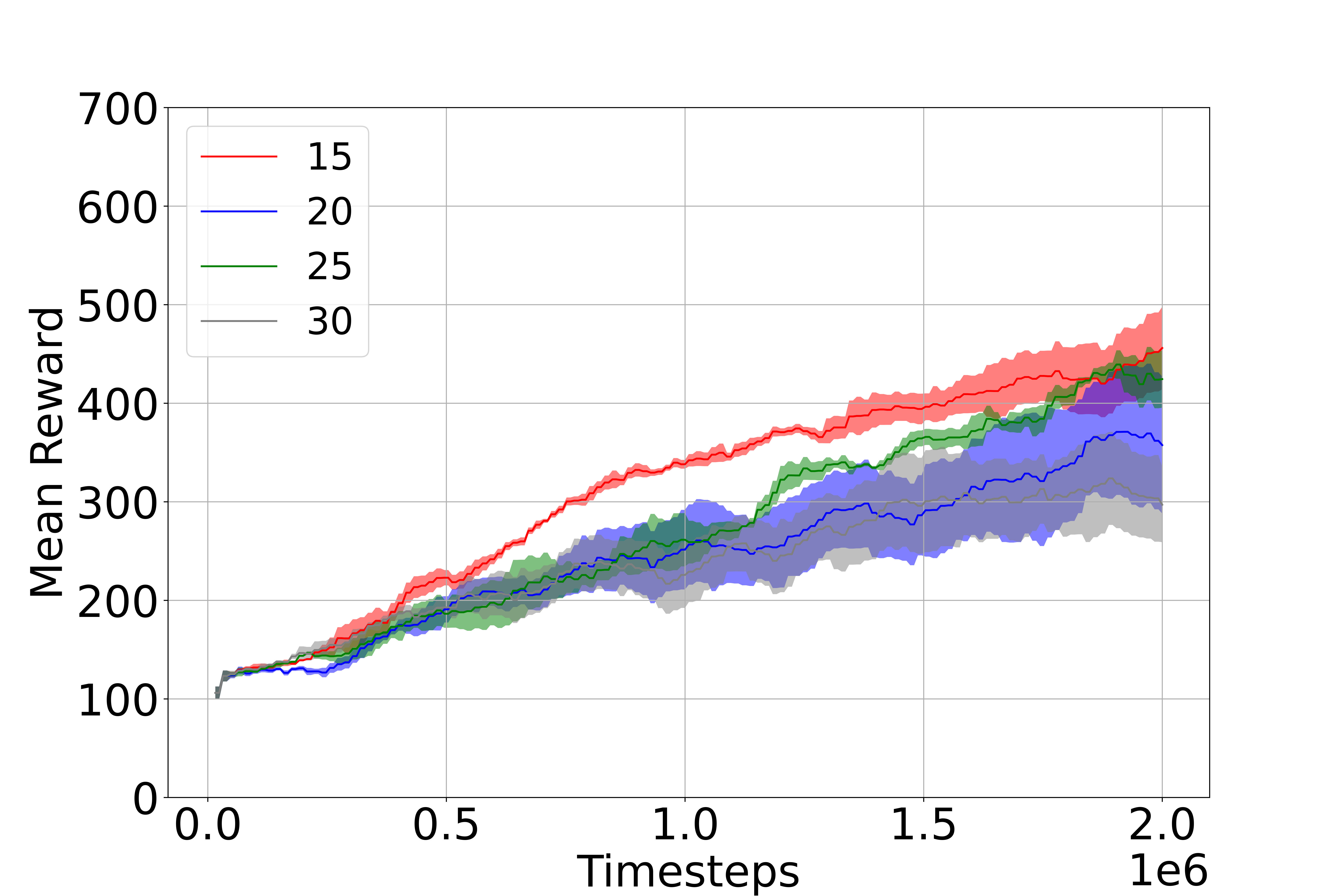}
        \caption{\eqref{eq:OtherQ} with $n=6$}
        \label{fig:quad_k_value}
   \end{subfigure}
       \caption{Episodic reward in Cheetah using different reward boundaries.}
       \label{fig:r_b_q_cheetah}
\end{figure}

For both Equation \ref{eq:Qfun_old} and \eqref{eq:OtherQ}, we consider a reward boundary for each class to be an equal part of some arbitrary maximum reward. In other words, if the maximum cumulative reward is $50$. If there are two rating classes, the boundary for the two rating classes is $25$. If there are three rating classes, the boundaries for the three rating classes are $50\times 1/3$ and $50\times 2/3$. The similar principle applies to the cases of more rating classes. We conduct the experiments using both \eqref{eq:Qfun_old} and \eqref{eq:OtherQ} across different reward boundaries (15, 20, 25, 30) for Cheetah when $n=2$ and $n=6$ over 5 runs. It can be seen from Figure \ref{fig:r_b_q_cheetah} that Equation \eqref{eq:Qfun_old} provides the best and most consistent performance is achieved when using a maximum reward of $25$ when $n = 2$ and $20$ when $n = 6$. However,  when using Equation \eqref{eq:OtherQ}, the best performance for $n=2$ and $6$ are consistently achieved when using a maximum reward of $15$. This suggests that different rating criteria can impact the performance, whereas the new function \eqref{eq:OtherQ} shows more stable performance across different reward boundaries overall.

\subsection{Adam vs AdamW}

\begin{figure}[ht!]
    \centering
    \begin{subfigure}[b]{0.49\columnwidth}
         \centering
         \includegraphics[width=\columnwidth]{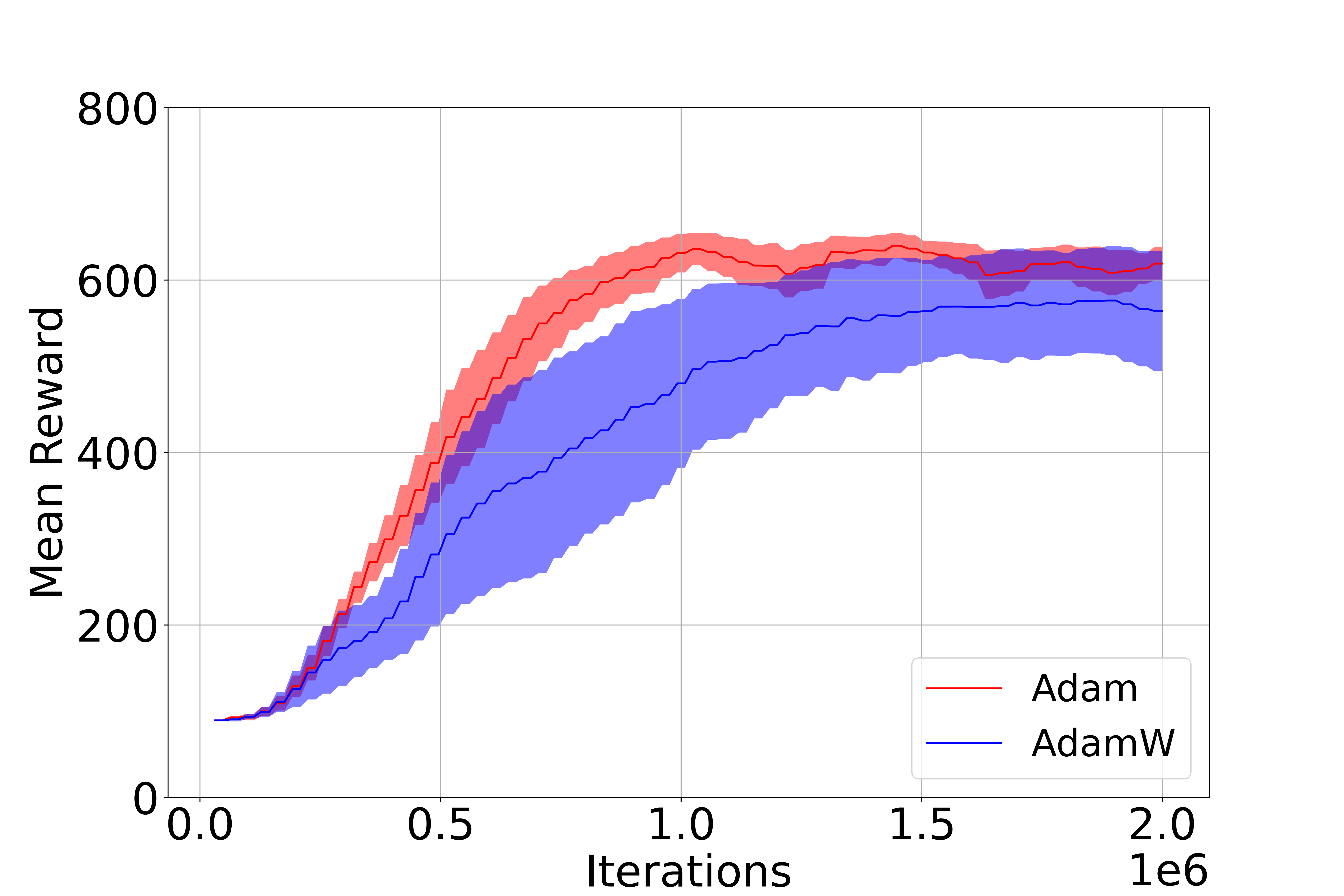}
         \caption{2 ratings}
         \label{fig:quad_k_value}
    \end{subfigure}
    \begin{subfigure}[b]{0.49\columnwidth}
         \centering
         \includegraphics[width=\columnwidth]{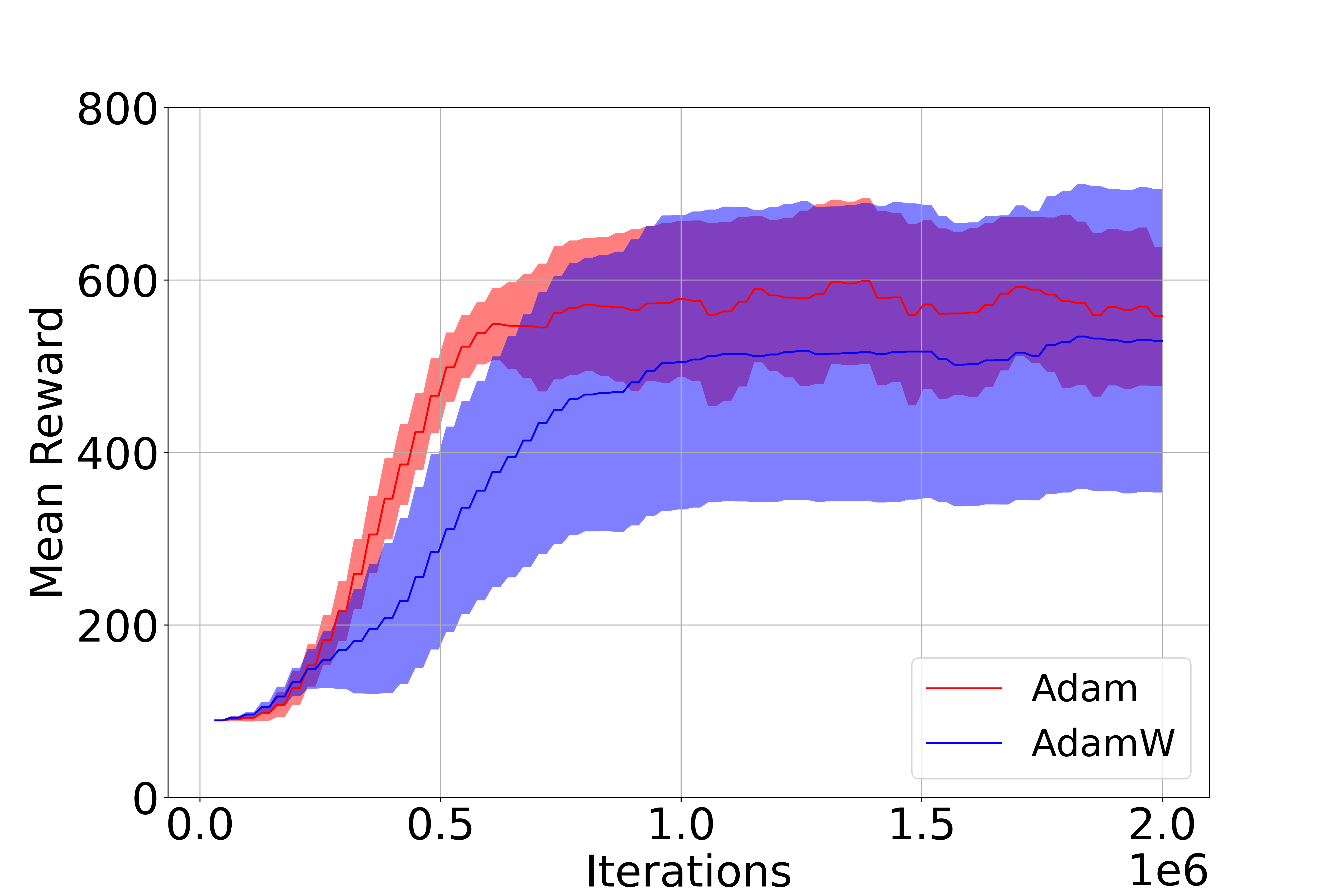}
         \caption{3 ratings}
         \label{fig:quad_k_value}
    \end{subfigure}
    \begin{subfigure}[b]{0.49\columnwidth}
         \centering
         \includegraphics[width=\columnwidth]{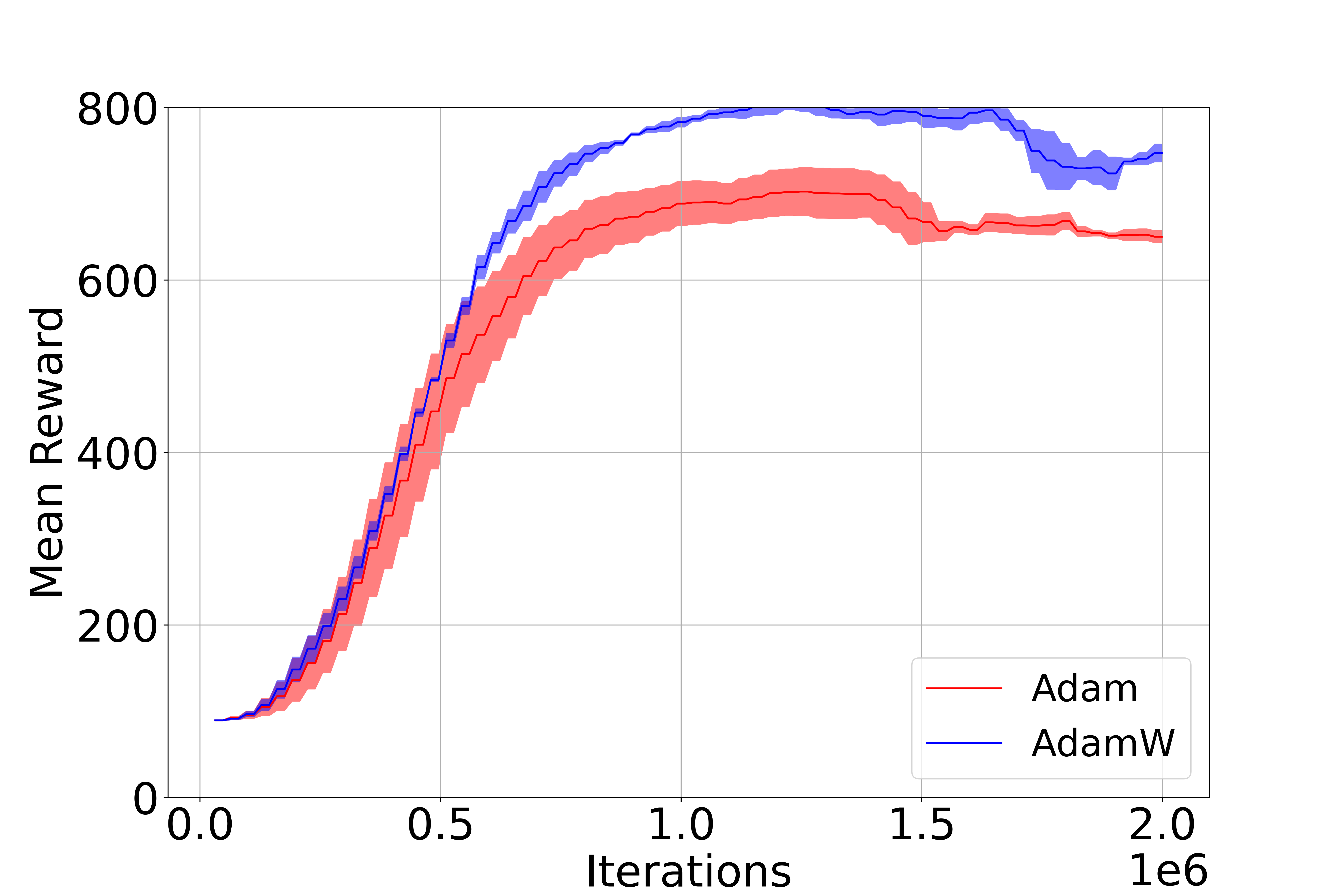}
         \caption{4 ratings}
         \label{fig:quad_k_value}
    \end{subfigure}
    \begin{subfigure}[b]{0.49\columnwidth}
         \centering
         \includegraphics[width=\columnwidth]{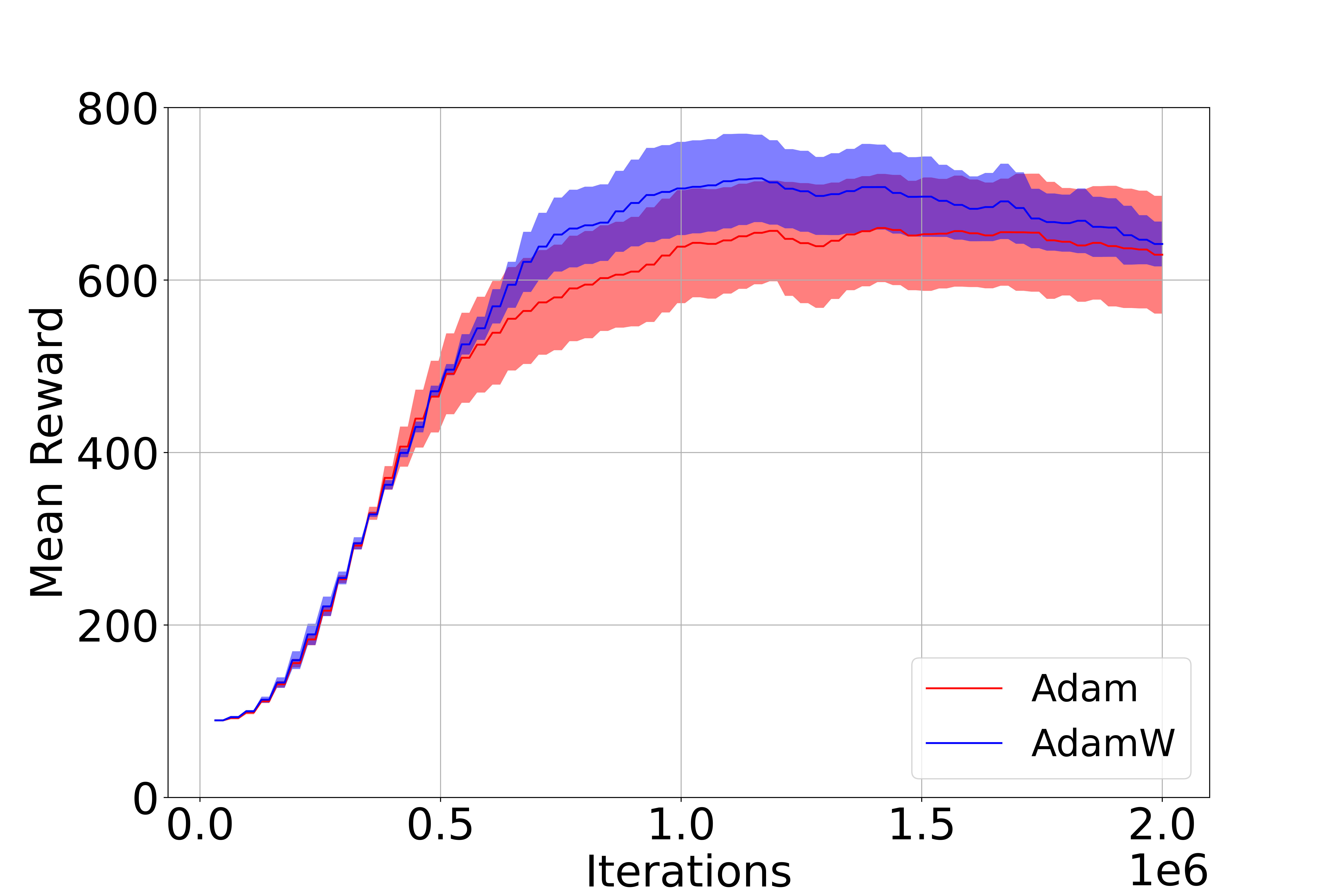}
         \caption{5 ratings}
         \label{fig:quad_k_value}
    \end{subfigure}
    \begin{subfigure}[b]{0.49\columnwidth}
         \centering
         \includegraphics[width=\columnwidth]{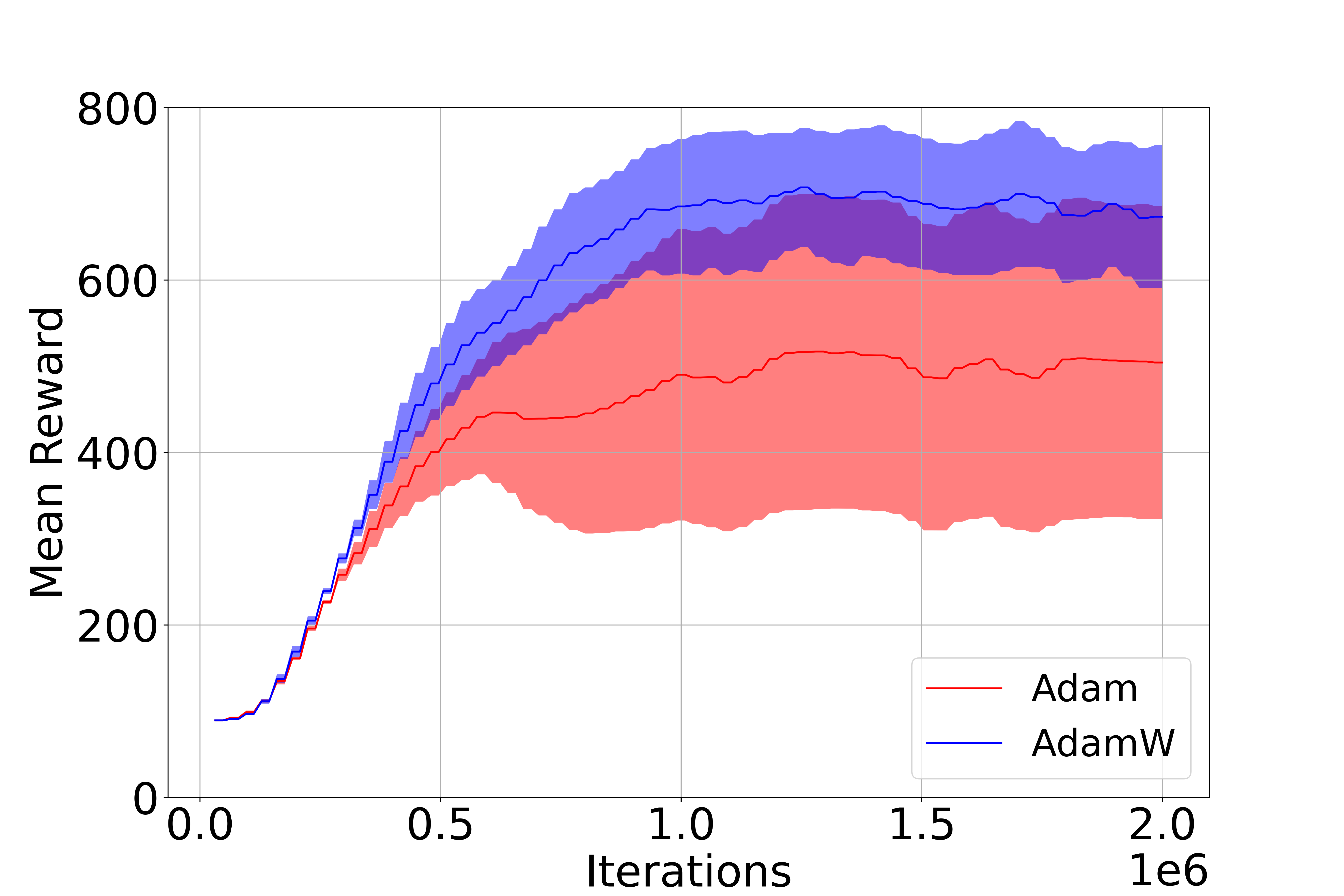}
         \caption{6 ratings}
         \label{fig:quad_k_value}
    \end{subfigure}
        \caption{Episodic reward using Adam and AdamW in Walker.}
        \label{fig:AdamW}
\end{figure}

This section investigates how the RbRL behaviors when changing the optimizer from Adam, used in the original code, to AdamW. Figure \ref{fig:AdamW} shows performance of both AdamW and Adam in Walker for 5 runs. It can be seen that both Adam and AdamW perform reasonably well. However, AdamW yields a higher overall performance than Adam. Additionally, Adam works better when the number of rating classes is small while AdamW works better when the number of rating classes is large. Finally, the use of AdamW produces higher consistency across different number of ratings classes. 

\subsection{Number of Hidden Layers}

\begin{figure}[ht!]
    \centering
    \begin{subfigure}[b]{0.49\columnwidth}
         \centering
         \includegraphics[width=\columnwidth]{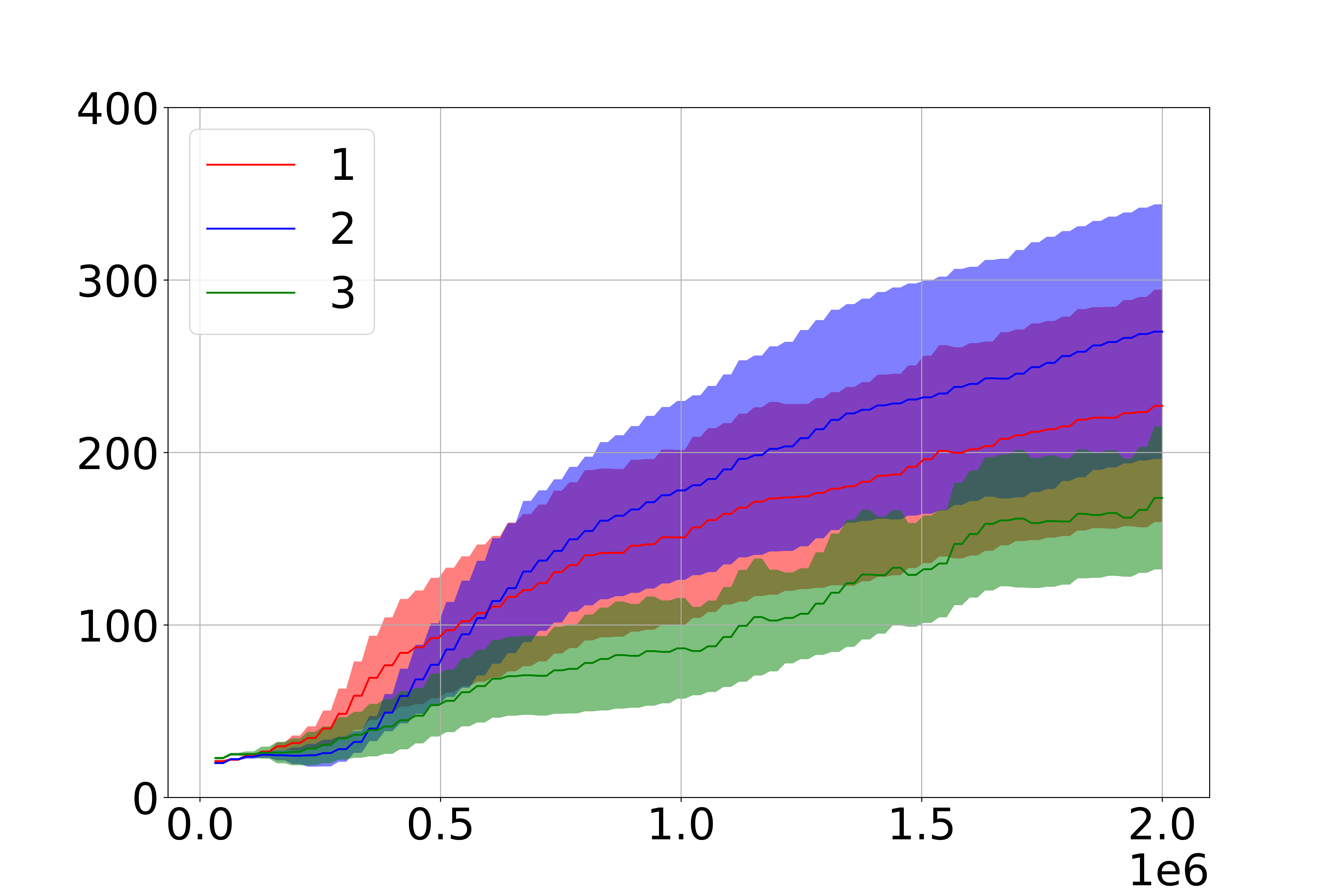}
         \caption{(3) with $n=2$}
         \label{fig:quad_k_value}
    \end{subfigure}
    \begin{subfigure}[b]{0.49\columnwidth}
         \centering
         \includegraphics[width=\columnwidth]{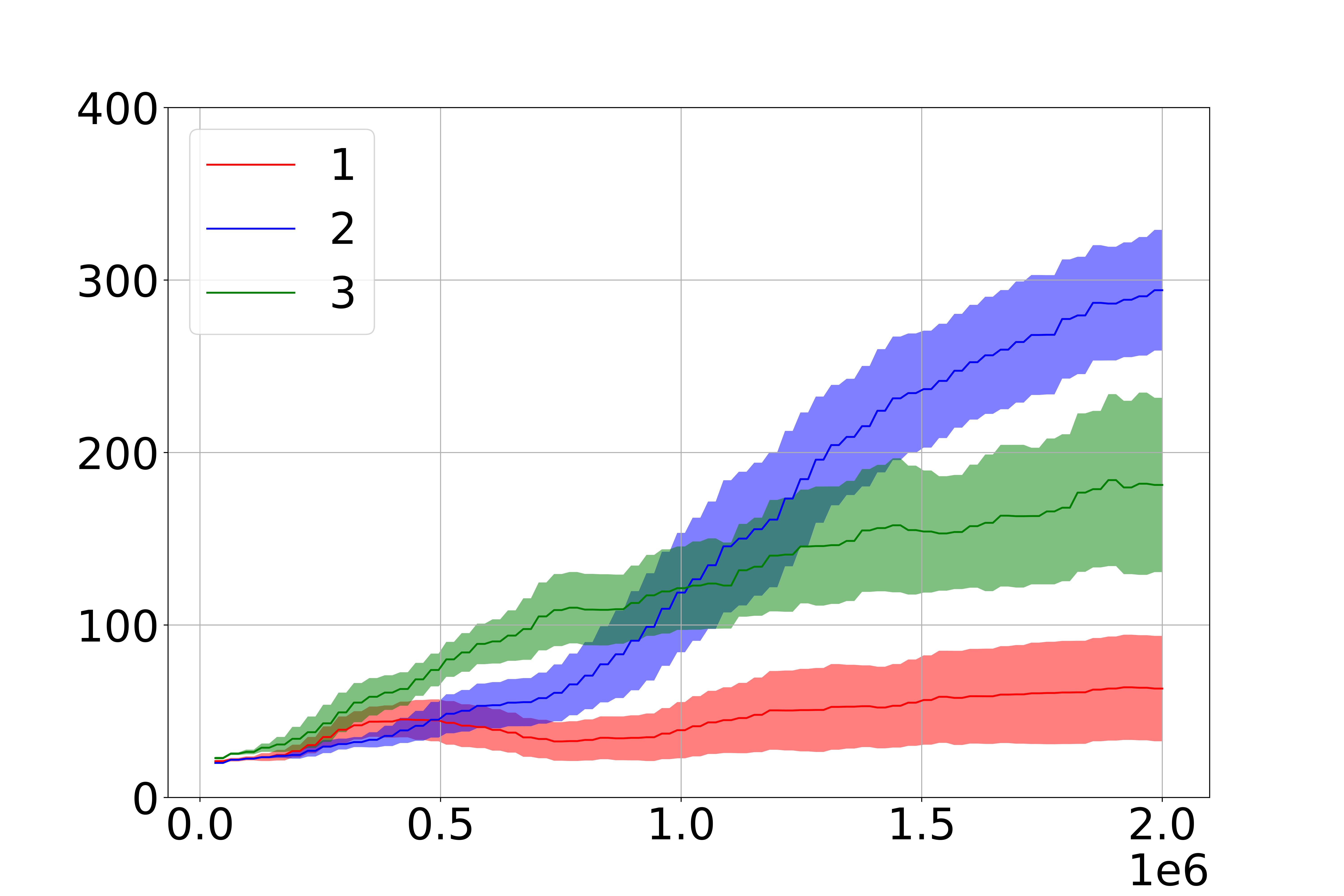}
         \caption{(5) with $n=2$}
         \label{fig:quad_k_value}
    \end{subfigure}
        \caption{Episodic reward in Cheetah using a different number of hidden layers.}
        \label{fig:Cheetah_hidden_layers}
\end{figure}

The size of the neural network plays a key role in the learning of reward models as overfitting or underfitting may occur when the size of the neural network is not selected appropriately. Here we investigate how the number of hidden layers may impact the performance of RbRL. Figure \ref{fig:Cheetah_hidden_layers} shows performance in Cheetah when changing the number of layers. It can be observed that RbRL performance and consistency of the model is best when using 2 layers for both equation \eqref{eq:Qfun_old} and \eqref{eq:OtherQ} . However, the use of 3 layers can achieve reasonable performance and consistency for both equation \eqref{eq:Qfun_old} and \eqref{eq:OtherQ}. We believe that the performance difference between using 2 layers and 3 layers is due to the model potentially overfitting. Further investigation into the performance on Walker and Quadruped as well as with different number of rating classes is currently being conducted to see what the best performance across these conditions is.

\subsection{Dropout}

Dropout has been shown to improve overall generalization within a model and helps to reduce the possibility of overfitting. We perform a set of experiments using $n=2$ and $n=6$ for both \eqref{eq:Qfun_old} and \eqref{eq:OtherQ} across 3 runs in the Quadruped, Cheetah and Walker. Table \ref{tab:Dropout} shows the mean and standard errors with different dropout values. It can be seen that for all cases other than Cheetah using Equation \eqref{eq:Qfun_old} and $n=2$ dropout improved the performance though there is no clear sign of what percent should be used as it is unique case by case. Additionally it should be noted that in some cases, if the dropout is not defined correctly, the performance can drop significantly. This leads us to believe that dropout will help alleviate the negative impact of overfitting in RbRL. However, further investigation needs to be done to see if there are any significant dropout values which generally perform better.



\begin{table*}[ht!]
  \centering
  \begin{tabular}{p{2cm}|p{2.5cm}|p{2.5cm}|p{2.5cm}|p{2.5cm}}
  \hline
    \multirow{2}{*}{Dropout in \%} & \multicolumn{4}{c}{Empirical Return for Walker } \\
    \cline{2-5}
    & RbRL (n=2) using \eqref{eq:Qfun_old} & RbRL (n=2) using \eqref{eq:OtherQ} & RbRL (n=6) using \eqref{eq:Qfun_old} & RbRL (n=6) using \eqref{eq:OtherQ}\\
    \hline
    $0$  & $564.04 \pm 70.02$         & $455.26 \pm 26.38$ & $673.41 \pm 82.78$          & $620.33 \pm 33.38$ \\
    $5$  & $614.3  \pm 39.17$         & $457.77 \pm 53.71$ & $\mathbf{849.51 \pm 19.24}$ & $792.43 \pm 29.42$ \\
    $10$ & $456.18 \pm 75.65$         & $564.14 \pm 19.51$ & $627.33 \pm 43.34$          & $795.3  \pm 30.84$ \\
    $20$ & $407.43 \pm 48.64$         & $\mathbf{598.78 \pm 58.37}$ & $687.16 \pm 90.96$          & $711.25 \pm 120.88$ \\
    $25$ & $\mathbf{599.2 \pm 30.34}$ & $555.23 \pm 43.74$ & $814.24 \pm 28.13$          & $\mathbf{795.59 \pm 20.92}$ \\
    $30$ & $557.77 \pm 53.71$         & $530.41 \pm 39.44$ & $733.3  \pm 46.21$          & $736.7  \pm 62.66$ \\
    \hline
    \multirow{2}{*}{Dropout in \%} & \multicolumn{4}{c}{Empirical Return for Quadruped } \\
    \cline{2-5}
    & RbRL (n=2) using \eqref{eq:Qfun_old} & RbRL (n=2) using \eqref{eq:OtherQ} & RbRL (n=6) using \eqref{eq:Qfun_old} & RbRL (n=6) using \eqref{eq:OtherQ}\\
    \hline
    $0$  & $275.25 \pm 59.83$   & $349.3 \pm 179.82$  & $453.6 \pm 79.06$   & $423.14 \pm 61.91$  \\
    $5$  & $205.43 \pm 33.5$    & $\mathbf{481.69 \pm 184.32}$ & $\mathbf{551.06 \pm 120.71}$ & $520.69 \pm 111.41$ \\
    $10$ & $172.86 \pm 46.42$   & $283.88 \pm 104.12$ & $323.15 \pm 51.03$  & $558.68 \pm 16.48$ \\
    $20$ & $162.8  \pm 8.24$    & $338.67 \pm 146.21$ & $323.14 \pm 121.88$ & $625.43 \pm 68.75$ \\
    $25$ & $160.53 \pm 3.37$    & $405.13 \pm 184.23$ & $321.22 \pm 69.23$  & $\mathbf{667.57 \pm 125.51}$ \\
    $30$ & $\mathbf{349.99 \pm 153.45}$  & $103.21 \pm 103.21$ & $493.92 \pm 57.48$  & $525.49 \pm 95.06$ \\
    \hline
    \multirow{2}{*}{Dropout in \%} & \multicolumn{4}{c}{Empirical Return for Cheetah } \\
    \cline{2-5}
    & RbRL (n=2) using \eqref{eq:Qfun_old} & RbRL (n=2) using \eqref{eq:OtherQ} & RbRL (n=6) using \eqref{eq:Qfun_old} & RbRL (n=6) using \eqref{eq:OtherQ}\\
    \hline
    $0$  & $\mathbf{583.95 \pm 30.46}$ & $363.37 \pm 4.62$  & $343.3 \pm 58.92$  & $397.5 \pm 72.49$  \\
    $5$  & $473.57 \pm 59.41$ & $427.58 \pm 29.37$ & $470.55 \pm 60.11$ & $393.52 \pm 23.56$ \\
    $10$ & $495.78 \pm 74.82$ & $460.62 \pm 36.61$ & $\mathbf{470.48 \pm 45.38}$ & $\mathbf{474.92 \pm 38.49}$ \\
    $20$ & $549.88  \pm 2.7$  & $\mathbf{525.27 \pm 53.31}$ & $470.13 \pm 15.55$ & $431.31 \pm 10.53$ \\
    $25$ & $511.22 \pm 54.88$ & $472.67 \pm 31.27$ & $425.48 \pm 26.96$ & $460.92 \pm 31.28$ \\
    $30$ & $465.89 \pm 26.71$ & $489.41 \pm 75.08$ & $441.05 \pm 22.61$ & $470.06 \pm 18.46$ \\
    \hline
\end{tabular}
\caption{Episodic reward for various dropout rates in Walker, Quadruped, and Cheetah.}
\label{tab:Dropout}
\end{table*}

\subsection{Activation Functions}

\begin{figure}[h!]
    \centering
    \begin{subfigure}[b]{0.49\columnwidth}
         \centering
         \includegraphics[width=\columnwidth]{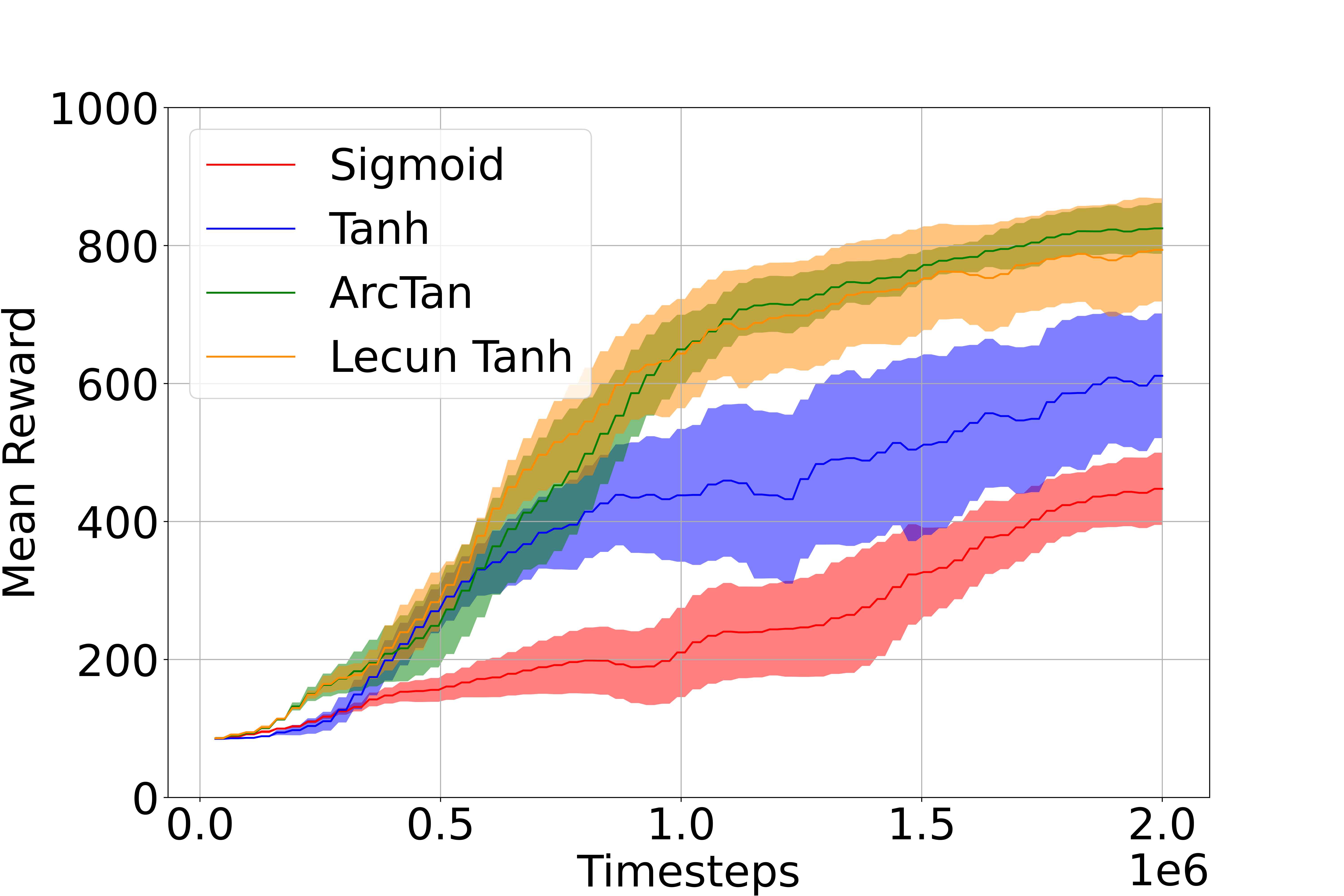}
         \caption{2 ratings}
         \label{fig:quad_k_value}
    \end{subfigure}
    \begin{subfigure}[b]{0.49\columnwidth}
         \centering
         \includegraphics[width=\columnwidth]{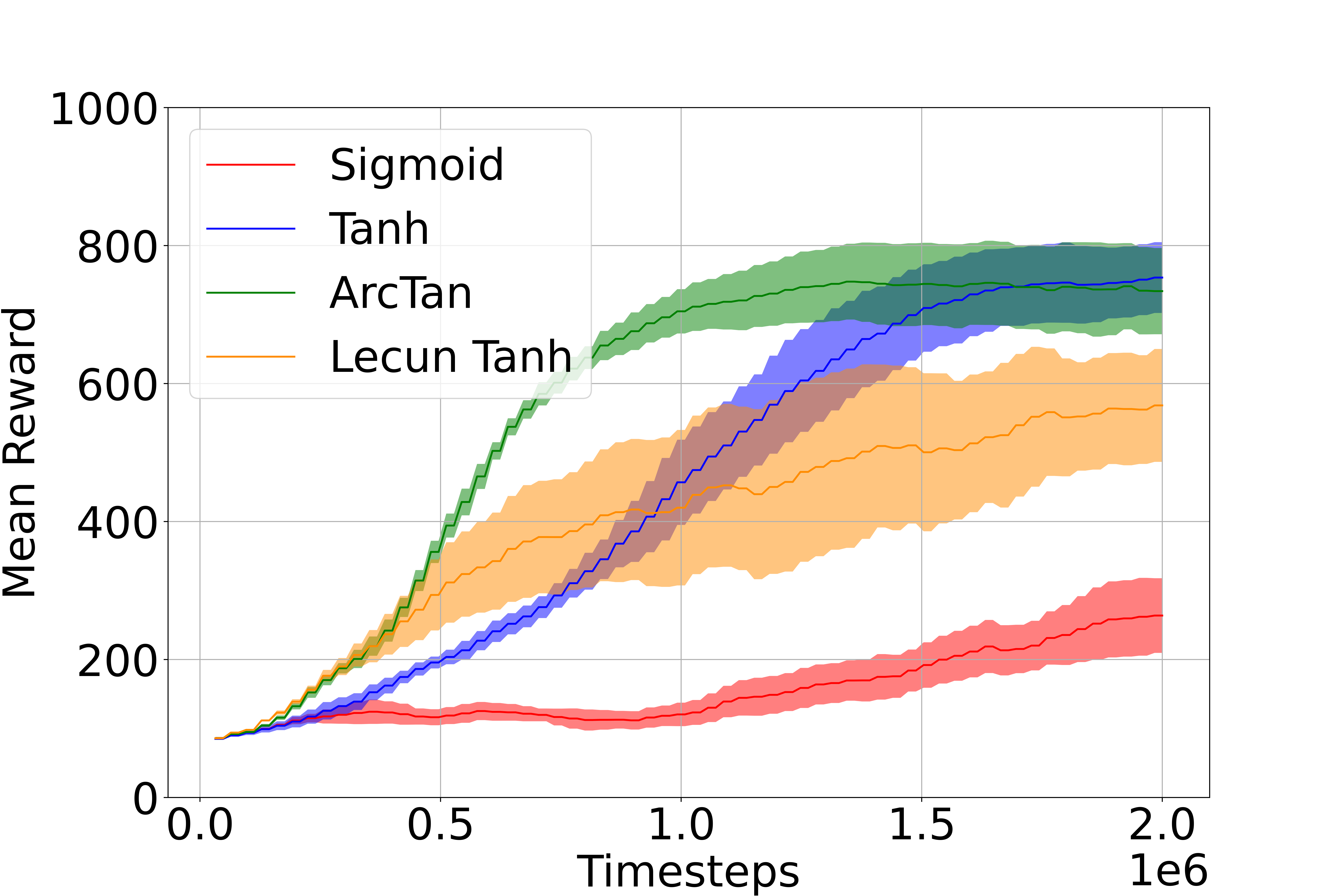}
         \caption{3 ratings}
         \label{fig:quad_k_value}
    \end{subfigure}
    \begin{subfigure}[b]{0.49\columnwidth}
         \centering
         \includegraphics[width=\columnwidth]{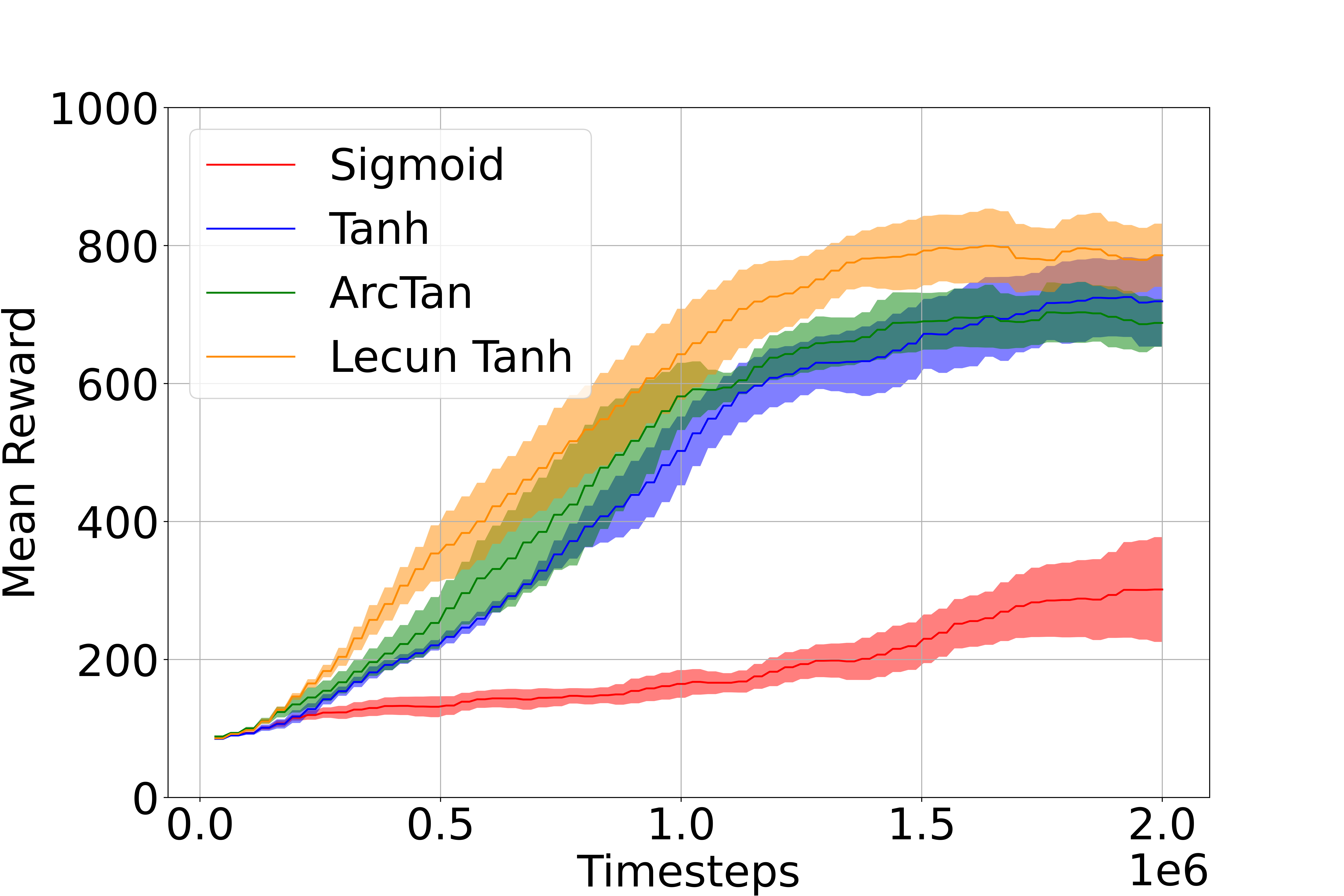}
         \caption{4 ratings}
         \label{fig:quad_k_value}
    \end{subfigure}
    \begin{subfigure}[b]{0.49\columnwidth}
         \centering
         \includegraphics[width=\columnwidth]{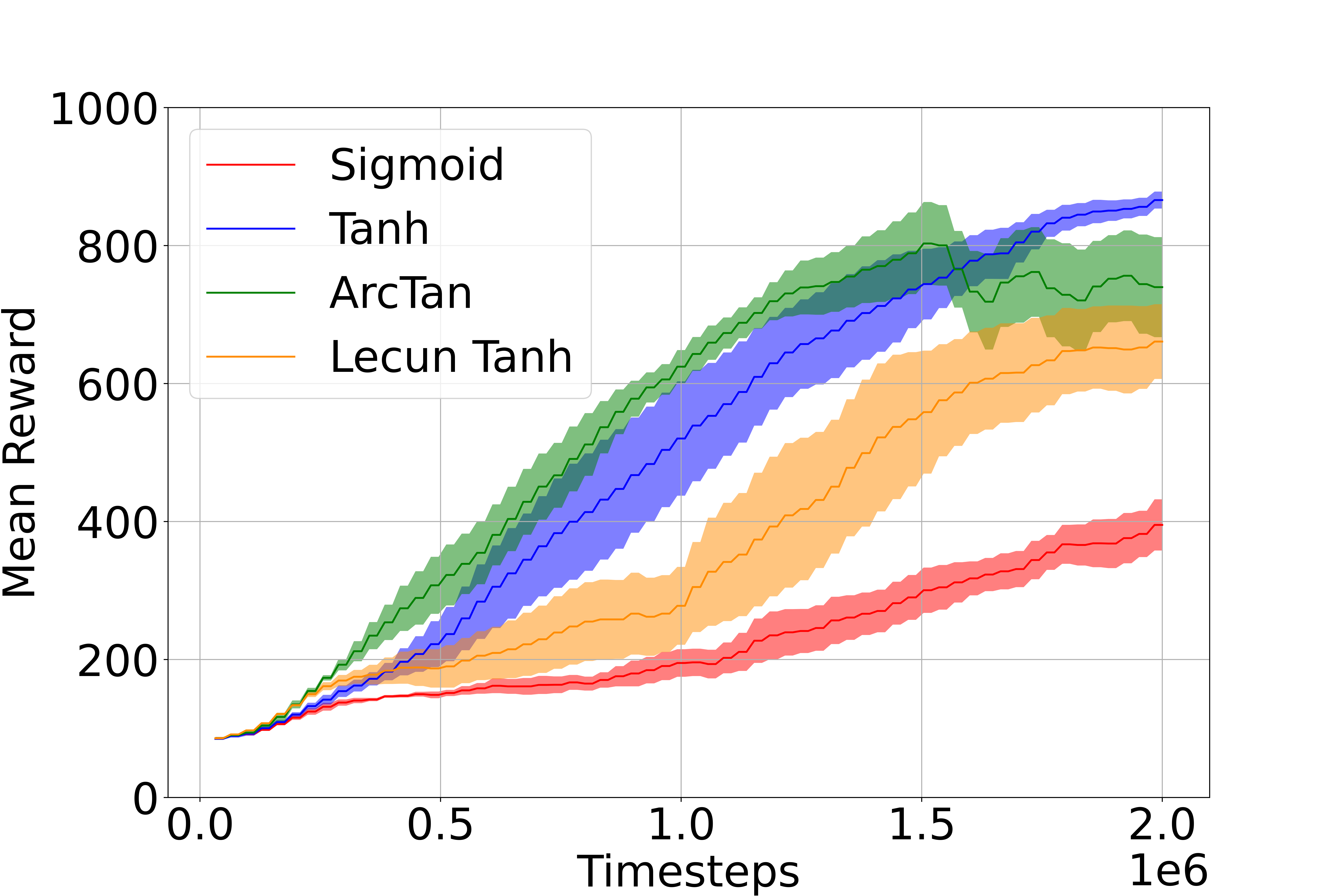}
         \caption{5 ratings}
         \label{fig:quad_k_value}
    \end{subfigure}
    \begin{subfigure}[b]{0.49\columnwidth}
         \centering
         \includegraphics[width=\columnwidth]{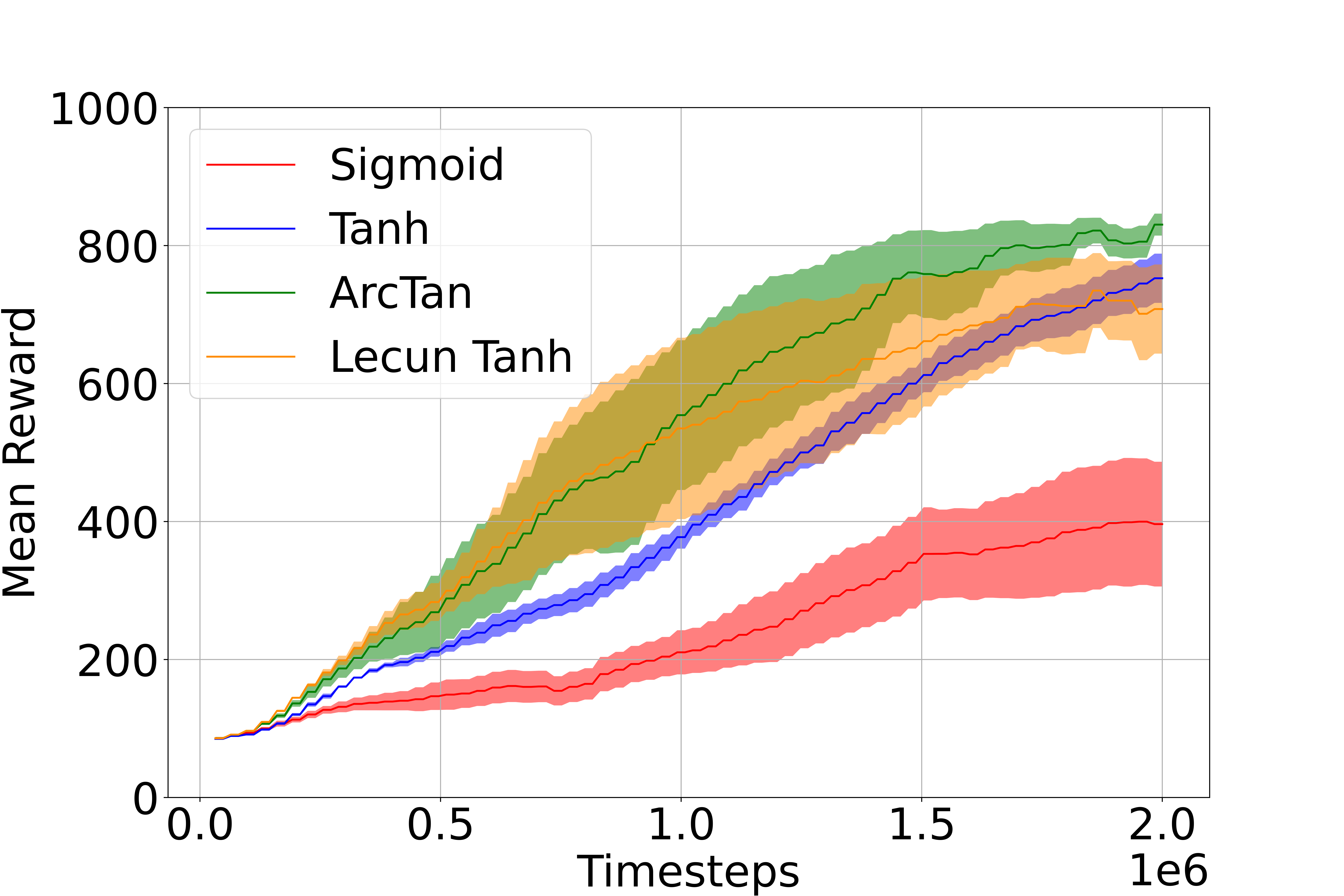}
         \caption{6 ratings}
         \label{fig:quad_k_value}
    \end{subfigure}
        \caption{Episodic reward in Walker using different activation functions.}
\label{fig:act_func}
\end{figure}

\begin{figure}[h!]
    \centering
    \begin{subfigure}[b]{0.49\columnwidth}
         \centering
         \includegraphics[width=\columnwidth]{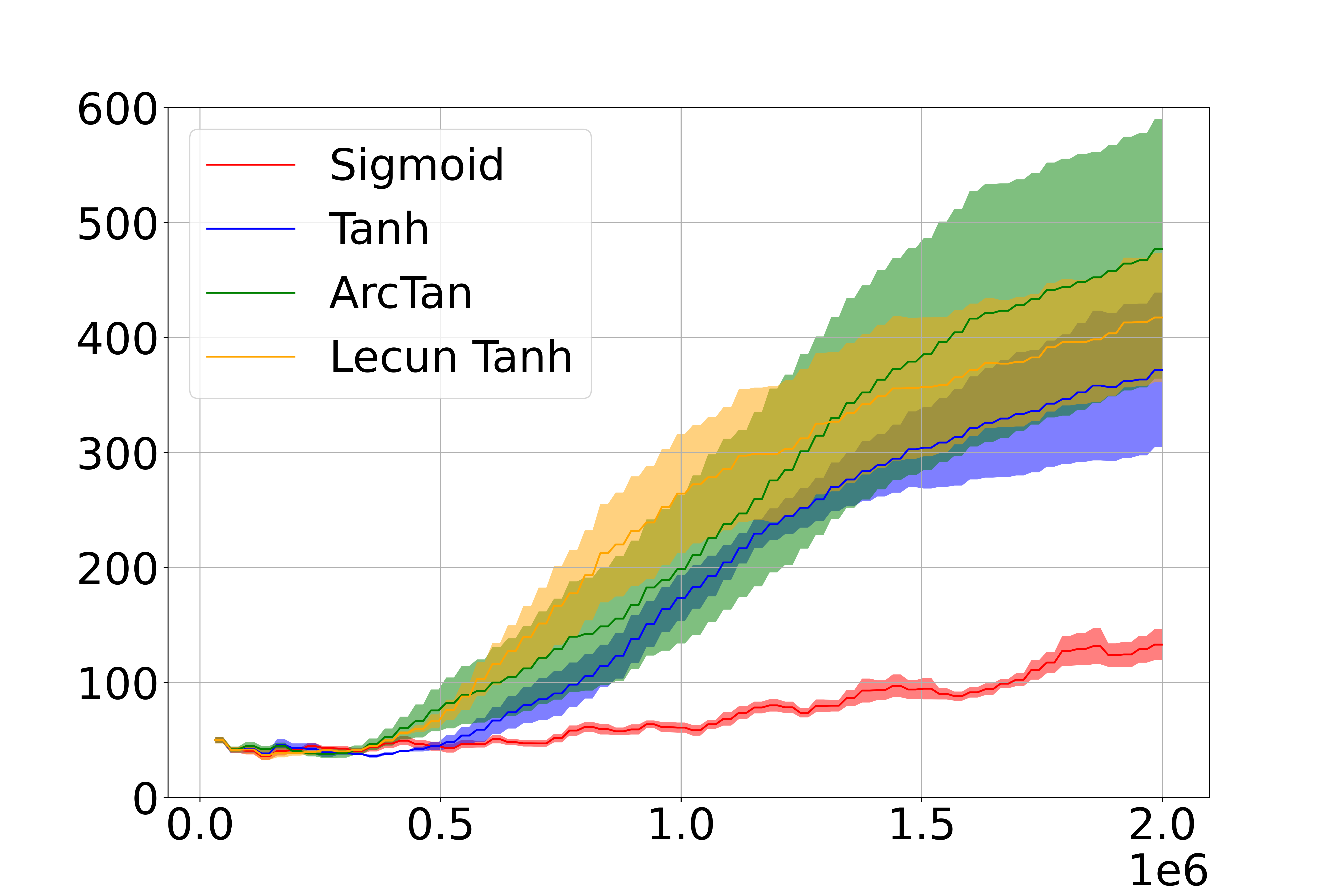}
         \caption{2 ratings}
         \label{fig:quad_k_value}
    \end{subfigure}
    \begin{subfigure}[b]{0.49\columnwidth}
         \centering
         \includegraphics[width=\columnwidth]{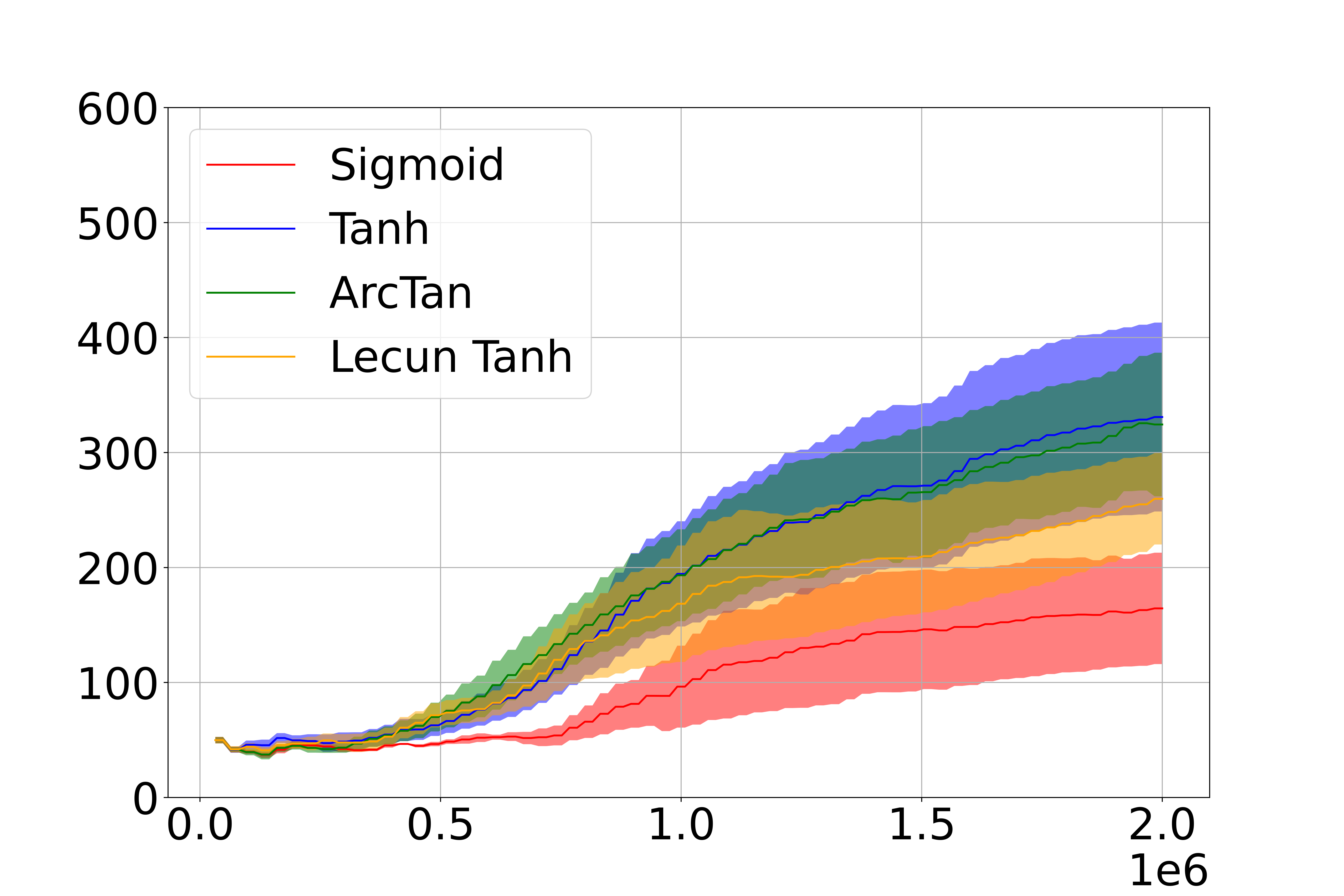}
         \caption{3 ratings}
         \label{fig:quad_k_value}
    \end{subfigure}
        \caption{Episodic reward in Quadruped using different activation functions.}
        \label{fig:quad_act}
\end{figure}

\begin{figure}[h!]
    \centering
    \begin{subfigure}[b]{0.49\columnwidth}
         \centering
         \includegraphics[width=\columnwidth]{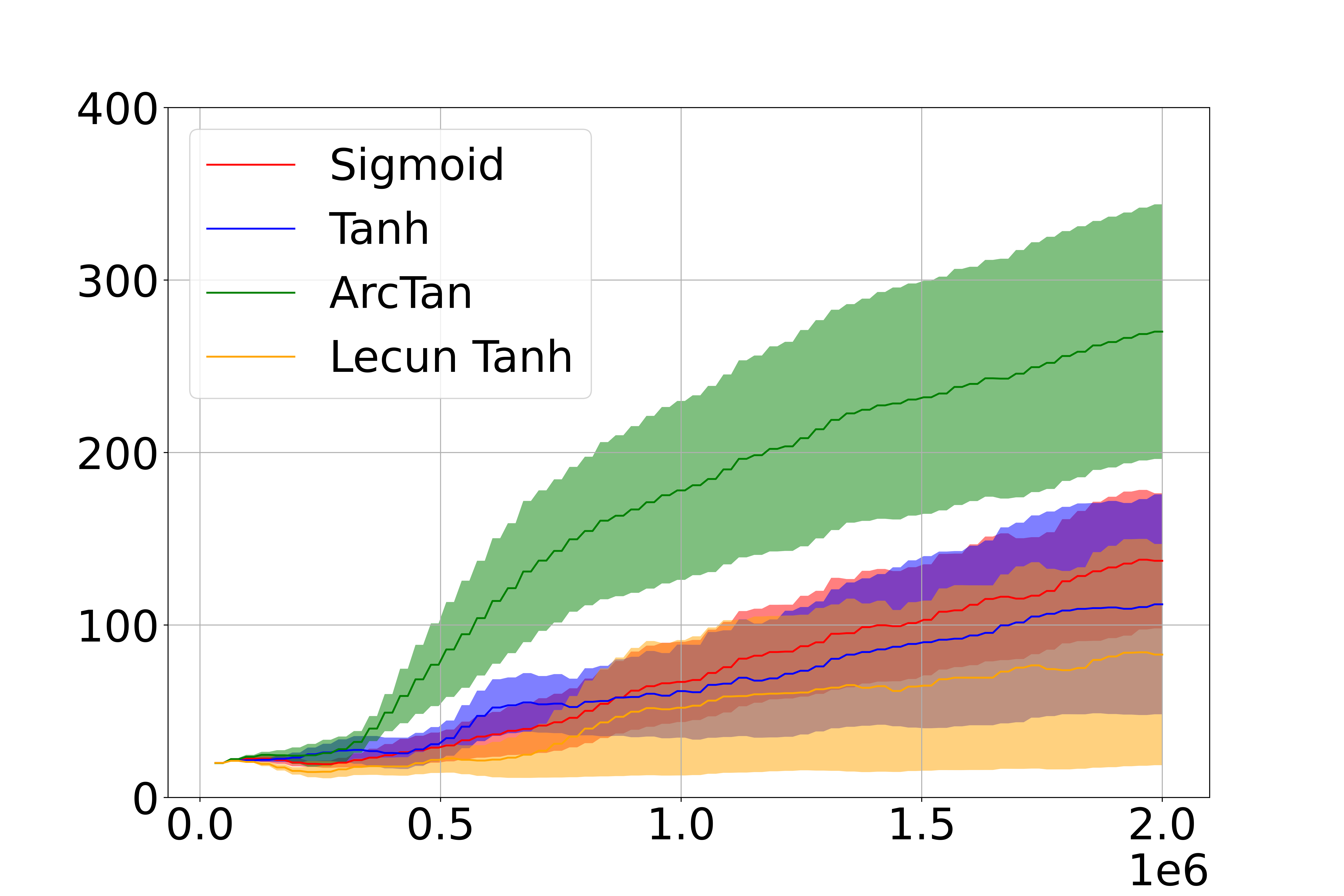}
         \caption{2 ratings}
         \label{fig:quad_k_value}
    \end{subfigure}
    \begin{subfigure}[b]{0.49\columnwidth}
         \centering
         \includegraphics[width=\columnwidth]{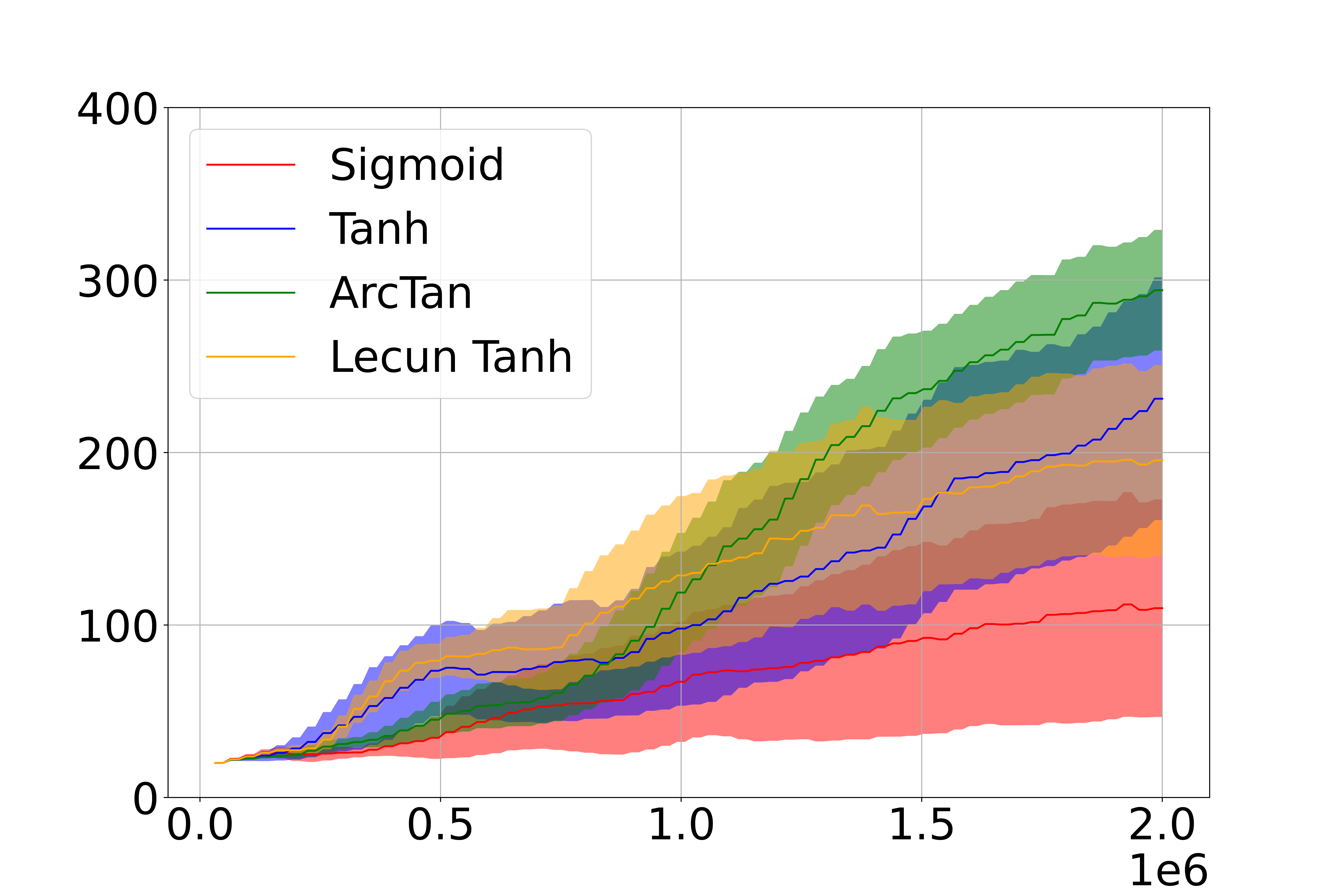}
         \caption{3 ratings}
         \label{fig:quad_k_value}
    \end{subfigure}
        \caption{Episodic reward in Cheetah using different activation functions.}
        \label{fig:cheetah_act}
\end{figure}

We investigate how various activation functions impact learning. Specifically we experiment with Sigmoid, Tanh, ArcTan, and Lecun Tanh to see if RbRL can yield different performance when using these different activation functions. We perform a set of experiments in Walker across 5 runs, shown in Figure \ref{fig:act_func}. It can be seen that the use of Tanh produces reasonable performance across different number of rating classes except for the case of $n=2$, this is what is used in the original paper \cite{white2024rating}. However, the use of ArcTan shows the best consistency and performance across all number of rating classes and environments. This is shown for Cheetah in Figure \ref{fig:cheetah_act}, where we can see for $n=2$ the performance increased from ~130 to almost 300 showing an improvement of over 100\%. As well, Quadruped showed better performance using ArcTan as seen in Figure \ref{fig:quad_act} where performance is better in $n=2$ and similar in $n=3$. Further experiments are being conducted to extend to higher number of rating classes. 

\subsection{Learning Rate}

We test the use of different learning rates to evaluate its impact on the consistency and overall performance for different number of rating classes for the Walker environment over 3 runs, shown in Table \ref{tab:lr}. It can be seen that the overall performance is best with a learning rate of $0.0005$. This learning rate can produce high overall performance and consistency altogether. 

\begin{table*}[ht!]
  \centering
  \begin{tabular}{p{2cm}|p{2.5cm}|p{2.5cm}|p{2.5cm}|p{2.5cm}|p{2.5cm}}
    \hline
    \multirow{2}{*}{Learning Rate} & \multicolumn{5}{c}{Empirical Return for Walker} \\
    \cline{2-6}
    & RbRL (n=2)  & RbRL (n=3) & RbRL (n=4)  & RbRL (n=5) & RbRL (n=6)\\
    \hline
    $ 0.0005$   & $\mathbf{638.68 \pm 19.64}$    & $\mathbf{602.2 \pm 49.58}$     & $\mathbf{835.59 \pm 9.41}$     & $769.38 \pm 5.37$  & $582.91 \pm 58.49$ \\
    $ 0.0003$   & $467.84 \pm 167.75$   & $504.58 \pm 40.33$    & $703.28 \pm 87.48$    & $626.13 \pm 45.61$ & $\mathbf{790.63 \pm 13.84}$ \\
    $ 0.0001$   & $550.44 \pm 31.06$    & $588.54 \pm 48.09$    & $605.5 \pm 41.88$     & $753.35 \pm 57.72$ & $779.34 \pm 45.75$ \\
    $ 0.00001$  & $355.99 \pm 153.23$   & $579.47 \pm 34.29$    & $522.23 \pm 122.14$   & $\mathbf{802.33 \pm 32.27}$ & $415.04 \pm 134.49$ \\
    \hline
\end{tabular}
\caption{Episodic reward for various learning rates in Walker.}
\label{tab:lr}
\end{table*}

\subsection{Optimized RbRL}

\begin{figure}[ht!]
    \centering
    \begin{subfigure}[b]{0.49\columnwidth}
         \centering
         \includegraphics[width=\columnwidth]{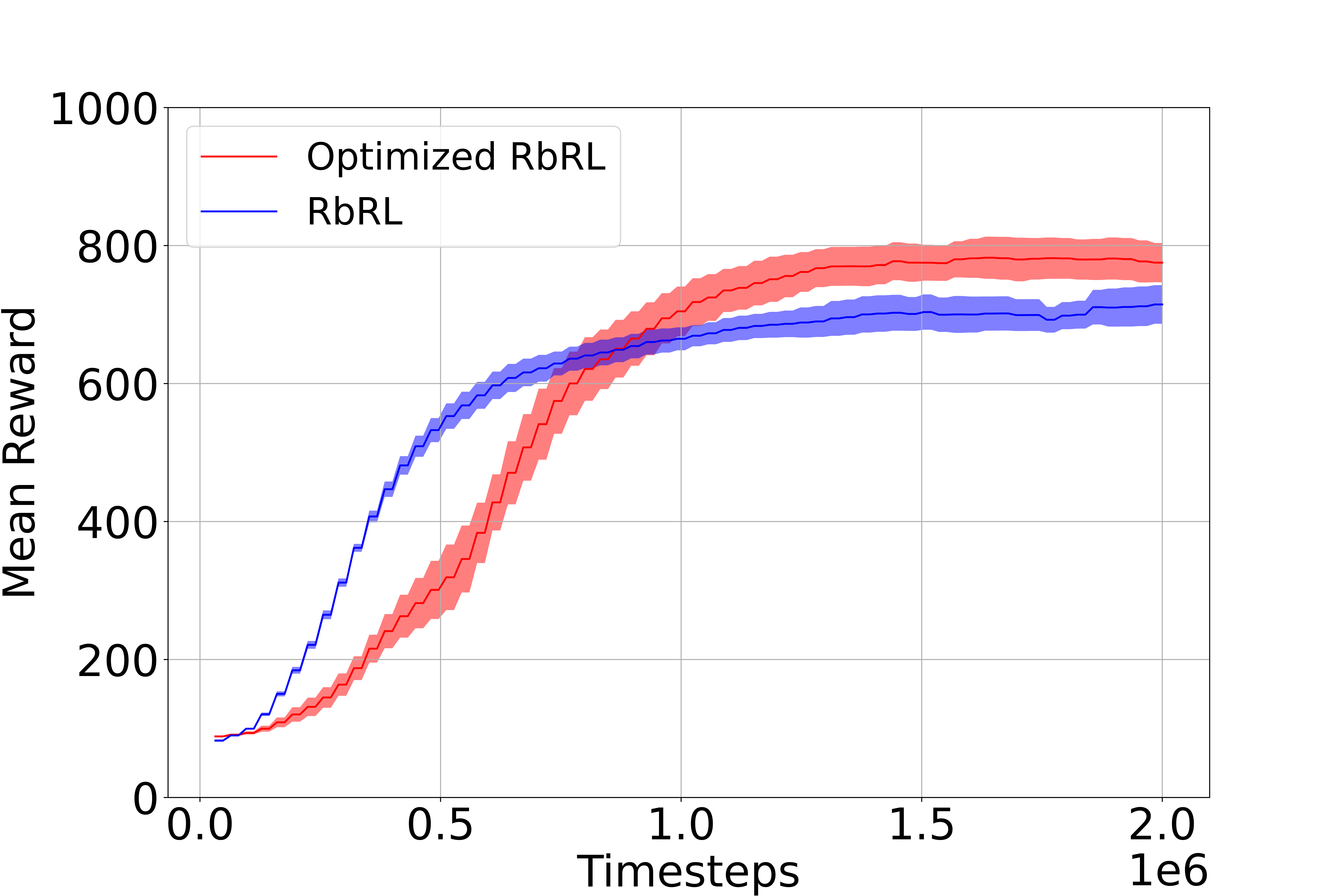}
         \caption{Walker}
         \label{fig:quad_k_value}
    \end{subfigure}
    \begin{subfigure}[b]{0.49\columnwidth}
         \centering
         \includegraphics[width=\columnwidth]{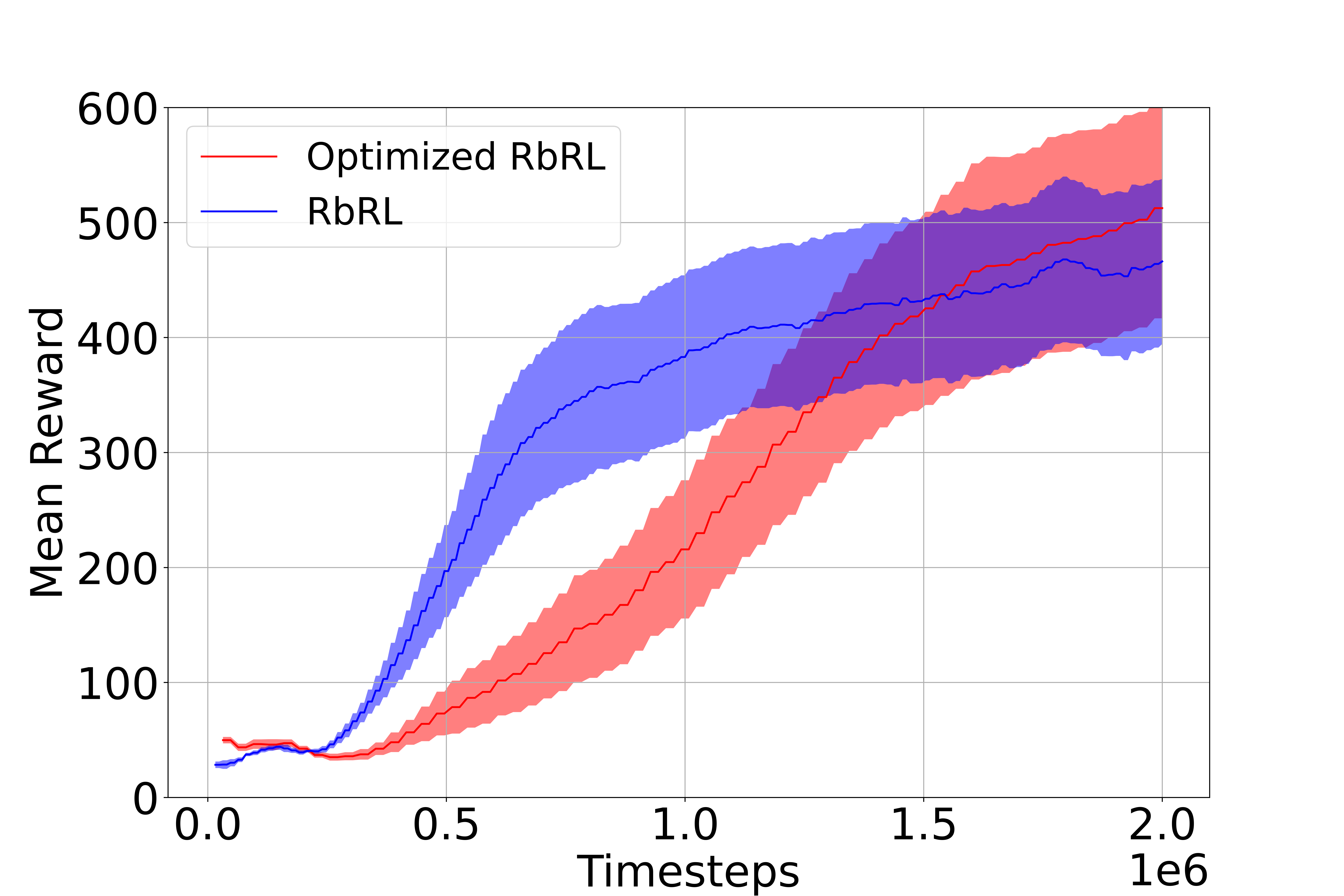}
         \caption{Quadruped}
         \label{fig:quad_k_value}
    \end{subfigure}
    \begin{subfigure}[b]{0.49\columnwidth}
         \centering
         \includegraphics[width=\columnwidth]{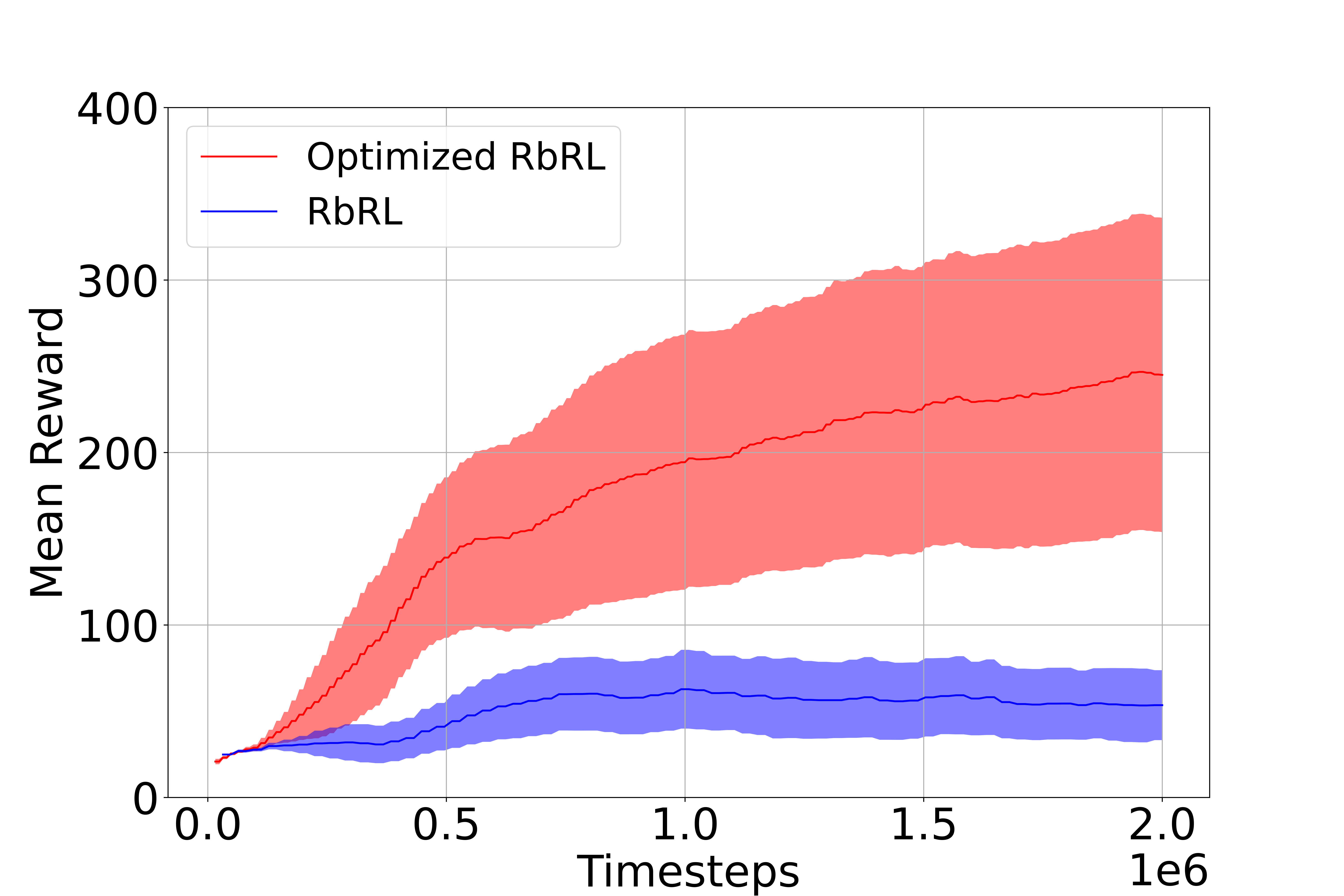}
         \caption{Cheetah}
         \label{fig:quad_k_value}
    \end{subfigure}
        \caption{Episodic reward using optimized parameters compared to original parameters in Walker, Quadruped and Cheetah.}
        \label{fig:diff_env}
\end{figure}

\begin{figure}[ht!]
    \centering
    \begin{subfigure}[b]{0.49\columnwidth}
         \centering
         \includegraphics[width=\columnwidth]{figures/walker1.png}
         \caption{2 ratings}
         \label{fig:quad_k_value}
    \end{subfigure}
    \begin{subfigure}[b]{0.49\columnwidth}
         \centering
         \includegraphics[width=\columnwidth]{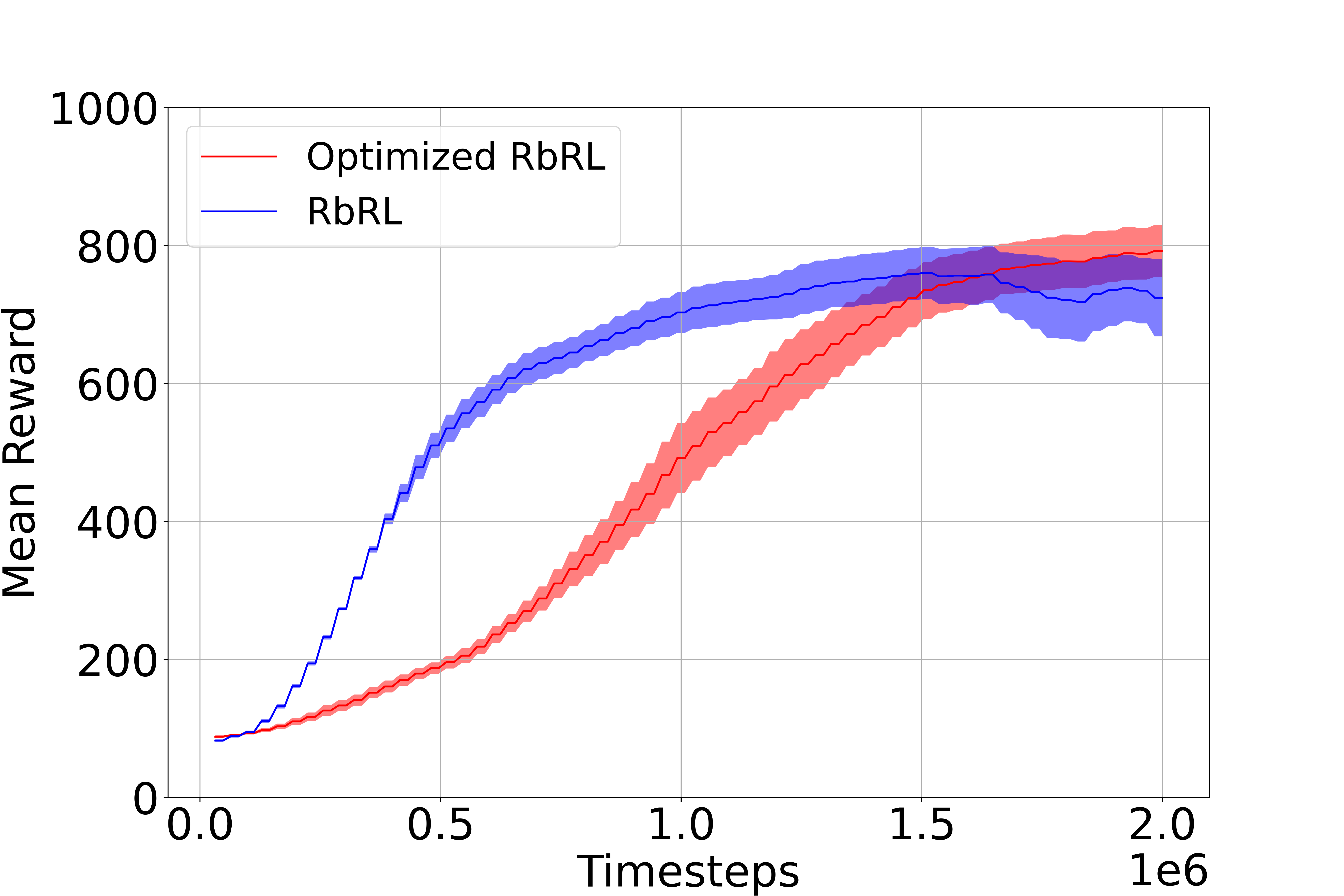}
         \caption{3 ratings}
         \label{fig:quad_k_value}
    \end{subfigure}
    \begin{subfigure}[b]{0.49\columnwidth}
         \centering
         \includegraphics[width=\columnwidth]{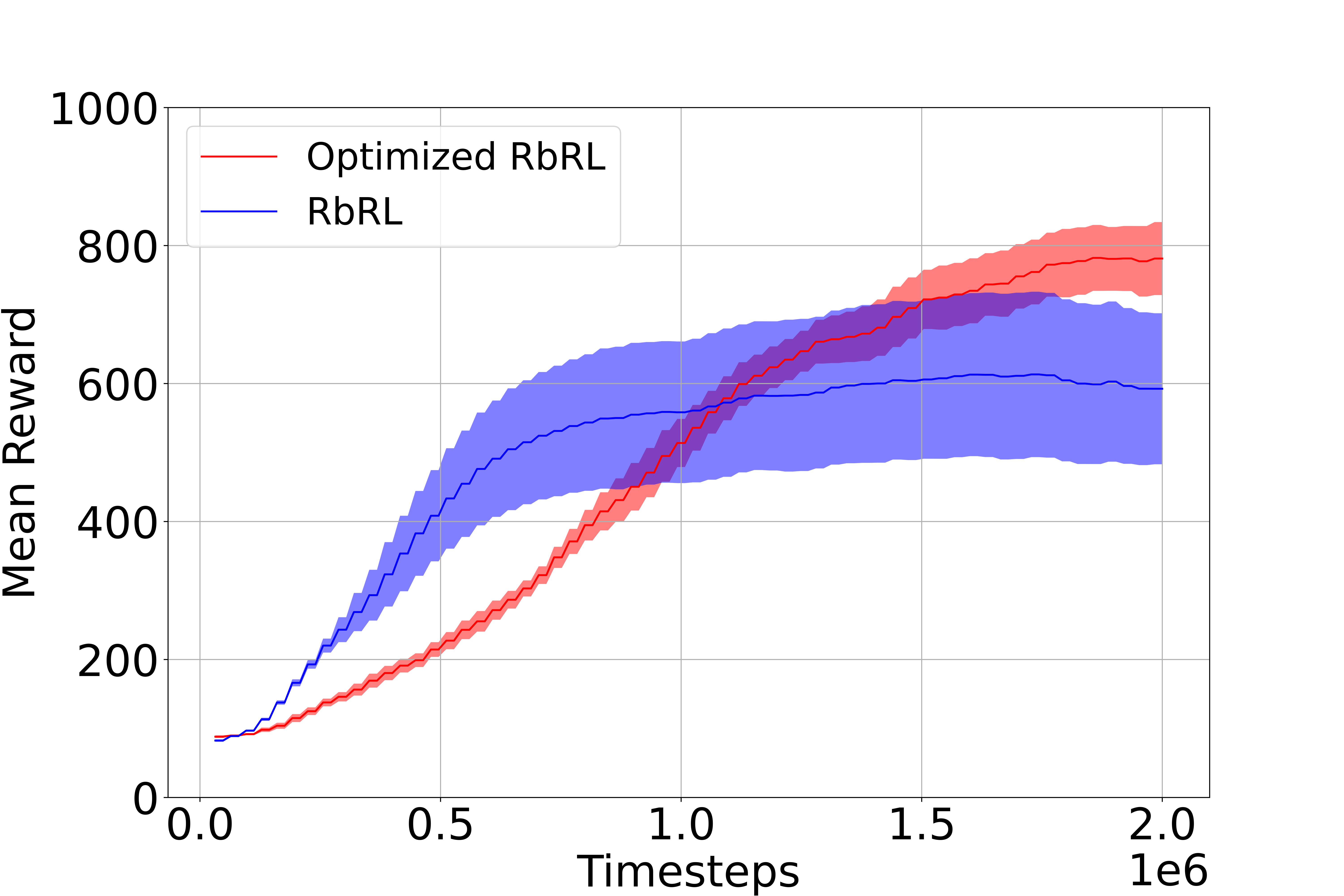}
         \caption{4 ratings}
         \label{fig:quad_k_value}
    \end{subfigure}
    \begin{subfigure}[b]{0.49\columnwidth}
         \centering
         \includegraphics[width=\columnwidth]{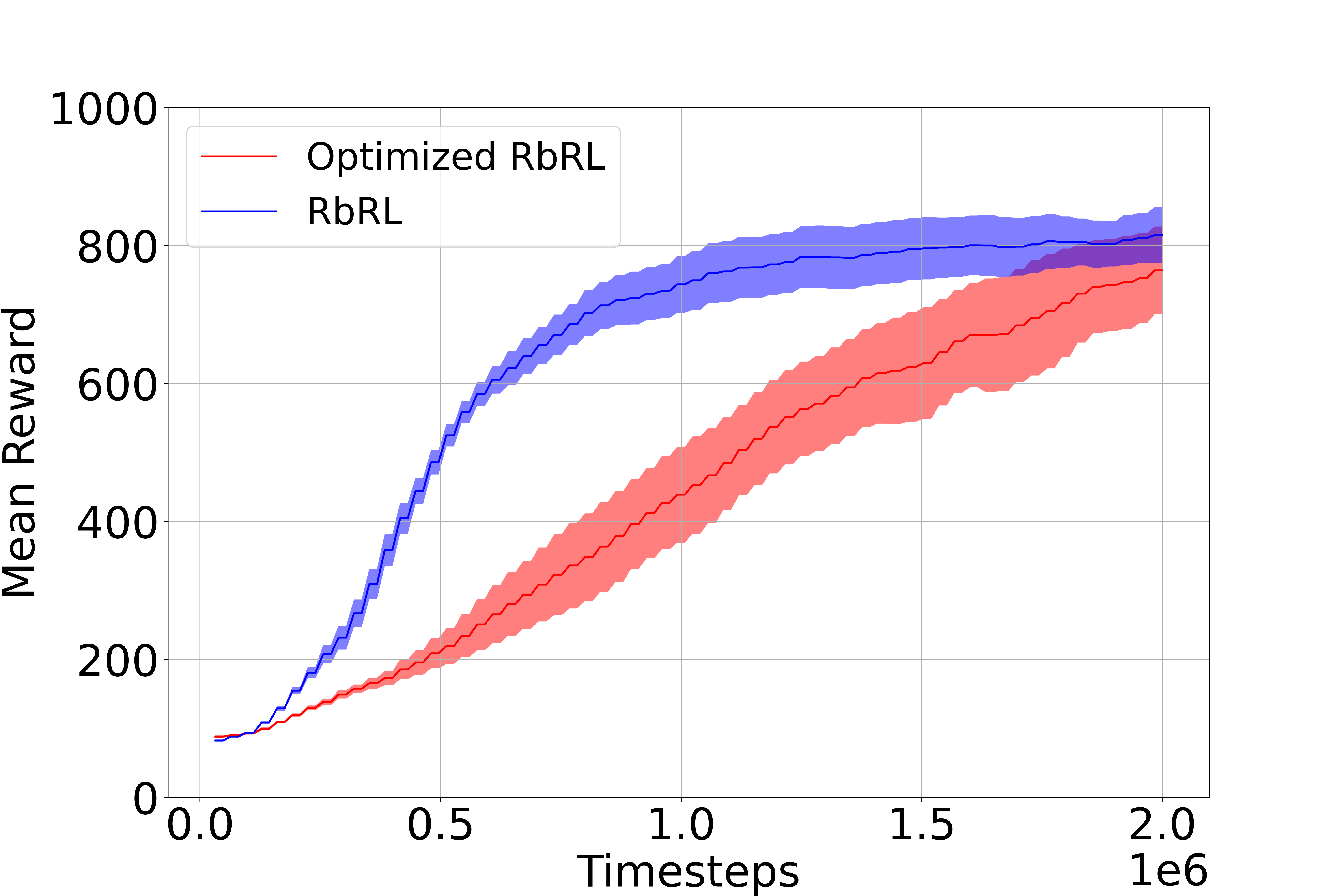}
         \caption{5 ratings}
         \label{fig:quad_k_value}
    \end{subfigure}
    \begin{subfigure}[b]{0.49\columnwidth}
         \centering
         \includegraphics[width=\columnwidth]{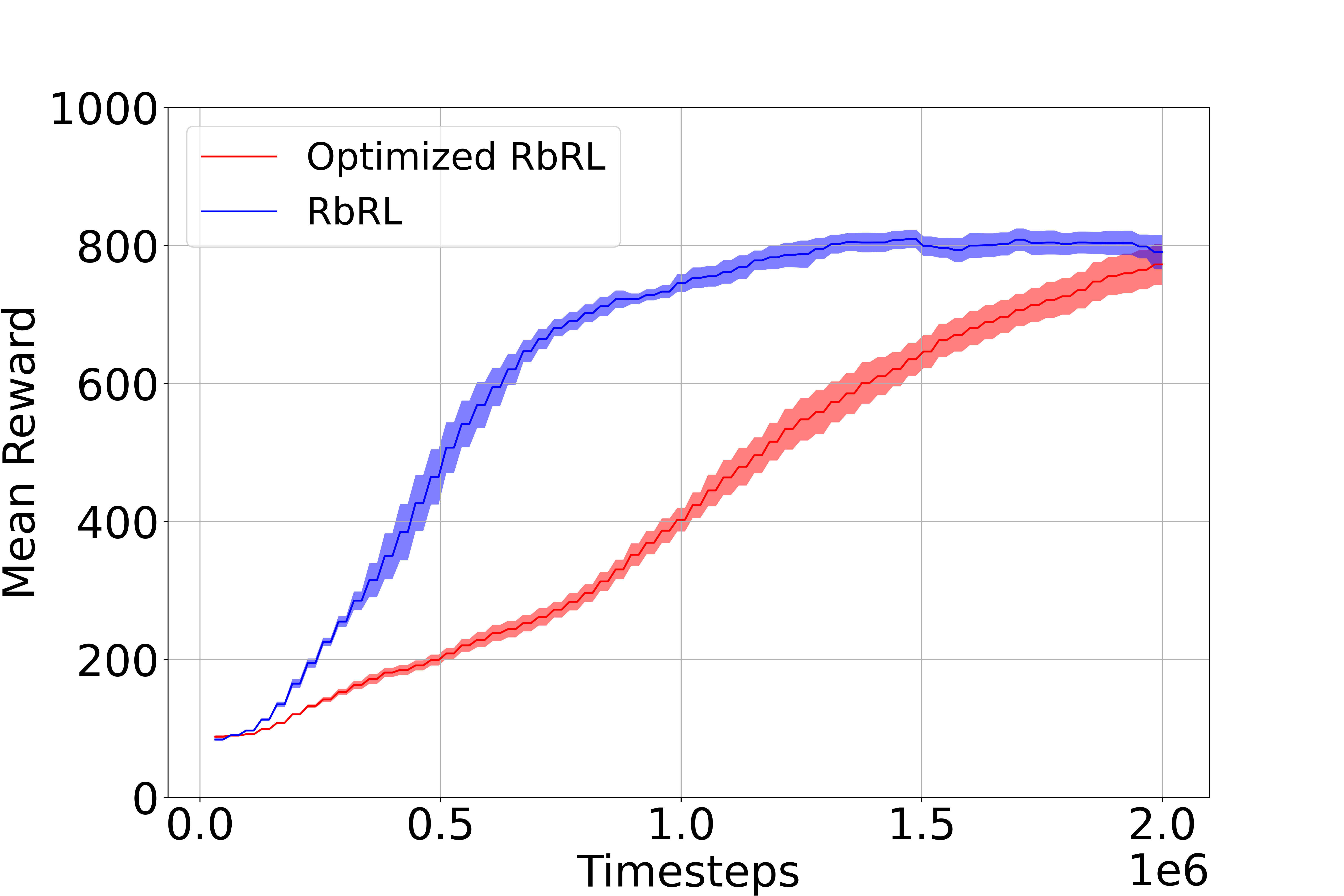}
         \caption{6 ratings}
         \label{fig:quad_k_value}
    \end{subfigure}
        \caption{Episodic reward using optimized parameters compared to original parameters in Walker across different number of rating classes.}
        \label{fig:diff_rating_class}
\end{figure}

We select the hyperparameters based on the best preforming and most consistent results across the three different environments. In particular, the confidence index is selected $k=1$. The optimizer is AdamW. 2 hidden layers are used with a dropout rate of 5\%. The activation function is ArcTan with a learning rate of 0.0005. 
\textcolor{black}{Figure \ref{fig:diff_env} illustrates the performance difference between optimized RbRL and RbRL for $n=2$ in Walker, Quadruped and Cheetah.} It can be observed that the optimized RbRL can show better performance and consistency in Walker and Quadruped and much better performance in Cheetah. Lastly, we investigate performance in Walker across different $n$ in Figure \ref{fig:diff_rating_class}. We can see that optimized RbRL can achieve better or the same performance as in traditional RbRL but with better consistency across 10 runs. This shows the impact of optimizing the hyperparameters in RbRL. More experiments are in progress to provide more comprehensive assessment of the optimized RbRL in different environments. 

\section{Discussion, Limitations, and Future Work}

\textcolor{black}{In this paper, we have shown that RbRL developed in~\cite{white2024rating} has room to yield better performance by optimizing its hyperparameters. Currently, our investigation into optimizing performance of RbRL is still in progress. However, our preliminary results show potential of selecting optimal hyperparameters to achieve better and more consistent performance. 
Our work serves as an initial study towards developing optimal RL methods from human cognitive inspired learning strategies, such as RbRL discussed in this paper. Yet, the optimization of the developed approach using the classic machine learning optimization techniques has not been properly investigated. Some interesting open questions include: (1) Is there a method which can be used to mitigate performance harm done by mislabeled segments? (2) Do these optimization techniques transfer from traditional robotics tasks to language tasks?  (3) Can we leverage both Preference-based Reinforcement Learning (PbRL) and RbRL in the same learning schema to improve consistency in performance? (4) Conduct human subject tests to validate these methods for users with different backgrounds.}

\bibliography{aaai25}

\end{document}